\documentclass[runningheads]{llncs}

\usepackage{eccv}

\usepackage{eccvabbrv}

\usepackage{graphicx}
\usepackage{booktabs}

\usepackage{adjustbox}
\usepackage{bm}
\usepackage{multirow}
\usepackage{makecell}
\usepackage{multicol}
\usepackage{color}
\usepackage{colortbl}
\usepackage{tabularx}
\usepackage{amsmath}
\usepackage{amssymb}
\usepackage{pifont}
\usepackage{bbm}

\usepackage[accsupp]{axessibility}  

\usepackage[pagebackref,breaklinks,colorlinks,citecolor=eccvblue]{hyperref}

\usepackage{orcidlink}

\begin{document}

\title{\raisebox{-0.30cm}{\includegraphics[width=1.0cm]{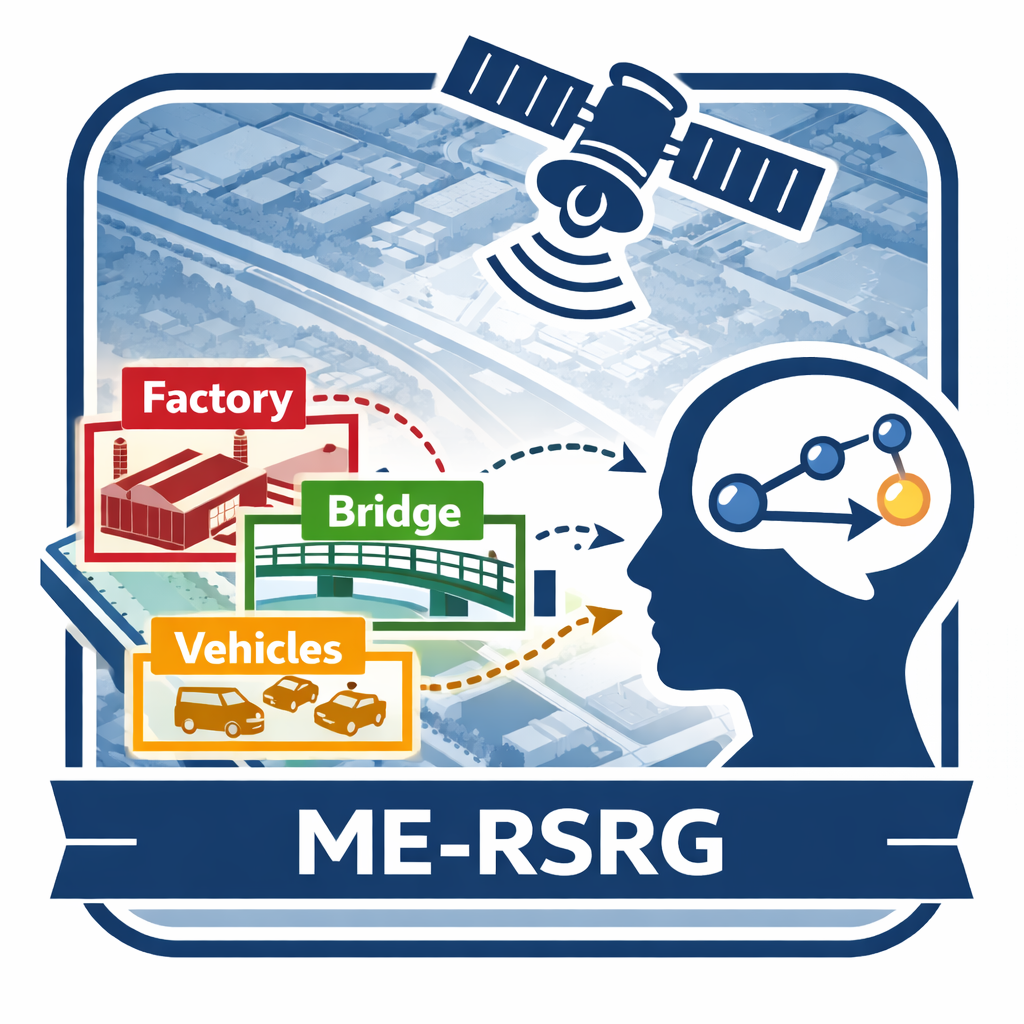}}\, Think and Answer ME: Benchmarking and Exploring Multi-Entity Reasoning Grounding in Remote Sensing}
\titlerunning{ME-RSRG}

\author{Shuchang Lyu\inst{1}\orcidlink{0000-0001-9769-7083} \and
Haiquan Wen\inst{2} \and Guangliang Cheng\inst{2}\dag \and
Meng Li\inst{1} \and Zheng Zhou\inst{1} \and You Zhou\inst{1} \and Dingding Yao\inst{3}, Zhenwei Shi\inst{1}}

\authorrunning{S.~Lyu et al.}

\institute{Beihang University, Beijing, China. \and University of Liverpool, Liverpool, UK. \\ \and
Institute of Acoustics, Chinese Academy of Sciences, Beijing, China \\ 
\dag \ Corresponding author: \email{Guangliang.Cheng@liverpool.ac.uk}}

\maketitle

\begin{abstract}
  Recent advances in reasoning language models and reinforcement learning with verifiable rewards have significantly enhanced multi-step reasoning capabilities. This progress motivates the extension of reasoning paradigms to remote sensing visual grounding task. However, existing remote sensing grounding methods remain largely confined to perception-level matching and single-entity formulations, limiting the role of explicit reasoning and inter-entity modeling. To address this challenge, we introduce a new benchmark dataset for Multi-Entity Reasoning Grounding in Remote Sensing (ME-RSRG). Based on ME-RSRG, we reformulate remote sensing grounding as a multi-entity reasoning task and propose an Entity-Aware Reasoning (EAR) framework built upon visual-linguistic foundation models. EAR generates structured reasoning traces and subject-object grounding outputs. It adopts supervised fine-tuning for cold-start initialization and is further optimized via entity-aware reward-driven Group Relative Policy Optimization (GRPO). Extensive experiments on ME-RSRG demonstrate the challenges of multi-entity reasoning and verify the effectiveness of our proposed EAR framework. Our dataset, code, and models will be available at \url{https://github.com/CV-ShuchangLyu/ME-RSRG}.
\end{abstract}

\section{Introduction}
\label{sec:intro}
\par Recently, reasoning language models such as OpenAI-o1~\cite{OpenAI-o1} and DeepSeek-R1~\cite{DeepSeek-R1} have shown strong multi-step reasoning capability by explicitly ``thinking'' before ``answering''. Meanwhile, reinforcement learning with verifiable rewards~\cite{Kimi,DeepSeek-R1,Visual-RFT,Reason-RFT} has emerged as a key mechanism for enhancing reasoning ability beyond supervised fine-tuning. 
\par Recent progress in reasoning models enables the extension of reasoning grounding to remote sensing, where scenes are characterized by large-scale spatial layouts, dense object distributions, and complex semantic relations. In such scenarios, adopting a language-centric paradigm allows grounding to move beyond perception-level matching toward structured reasoning with Chain-of-Thought (CoT). It makes remote sensing intelligence more interpretable.
\par Despite these advances, most existing remote sensing visual grounding (RSVG) methods~\cite{DIOR-RSVG,RSVG-HR,LGFormer,CSDNet} are still formulated as perception-level matching problems, focusing on aligning textual queries with visual regions. \textit{This paradigm typically considers a single target entity and places limited emphasis on inter-entity relations.} Consequently, incorporating intermediate reasoning steps in this setting may result in superficial reasoning that provides limited interpretability.
\par Specifically, this challenge can be further decomposed into two key issues. \textbf{(1) Existing remote sensing grounding datasets are limited to single-entity annotations.} It lacks structured supervision for multiple entities and their relations, which limits systematic exploration of multi-entity reasoning. \textbf{(2) Current grounding models are mainly designed under a matching-oriented paradigm.} As a result, supporting reasoning grounding in remote sensing remains challenging without explicit mechanisms for modeling entity roles, inter-entity relations, and intermediate reasoning processes.
\par To address the first issue, we construct a multi-entity reasoning grounding dataset, termed ME-RSRG. Unlike existing datasets that annotate a single target region per referring description, each sample in ME-RSRG explicitly involves a subject entity together with one or more object entities. By providing structured annotations for entity roles and language description, ME-RSRG establishes a unified foundation for multi-entity reasoning grounding in remote sensing.
\begin{figure}[!t]
  \centering
\includegraphics[width=1.0\textwidth]{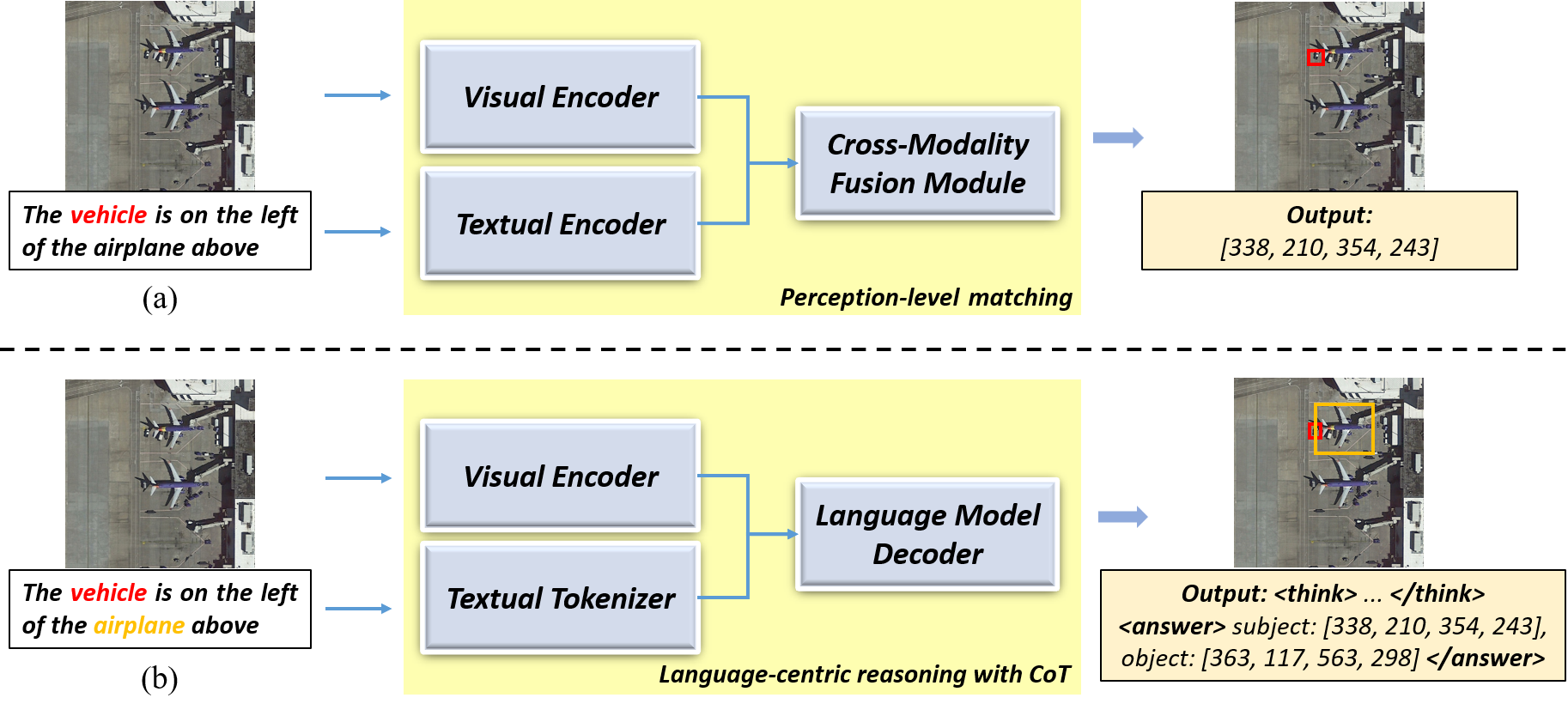}
  \caption{\textbf{Paradigm comparison between (a) matching-based visual grounding and (b) multi-entity reasoning grounding in remote sensing.}}
  \label{Fig1}
\end{figure}
\par To address the second issue, we propose a reasoning-oriented visual-linguistic framework, termed Entity-Aware Reasoning (EAR) framework based on ME-RSRG. As shown in Fig.~\ref{Fig1} (b), the visual encoder extracts and tokenizes visual representations, while the language input is processed as a unified sequence. These visual and linguistic tokens are then fed into a language model decoder, which serves as the core engine to perform reasoning. Instead of directly predicting a single target location, EAR generates structured outputs by explicitly producing a reasoning trace in the ``think'' form, followed by the subject-object grounding results in the ``answer'' form. To optimize EAR, we adopt a two-stage training strategy. Supervised fine-tuning (SFT) serves as a first stage cold start to initialize the model, enabling it to generate structured subject-object grounding outputs with coherent reasoning traces. Building upon this initialization, we further introduce Group Relative Policy Optimization (GRPO)~\cite{GRPO} for second stage optimization. It refines the model through entity-aware rewards that explicitly evaluate grounding correctness for different entity roles.
\par Fig.~\ref{Fig1} illustrates the paradigm comparison between matching-based visual grounding and multi-entity reasoning grounding in remote sensing. In the conventional RSVG paradigm, grounding is formulated as aligning a single referring expression with one target subject. In contrast, the proposed multi-entity reasoning grounding paradigm explicitly extracts multiple entities and produces structured reasoning-aware outputs for more interpretable understanding of complex remote sensing scenes. Moreover, our proposed multi-entity formulation is inherently more suitable for reasoning-oriented grounding rather than simple matching. In remote sensing imagery, entities are often distributed over large spatial extents and exhibit rich contextual dependencies, where the identity of a target is frequently determined by its relations to other entities rather than by isolated appearance cues. Meanwhile, several remote sensing methods~\cite{SegEarth-R1, RemoteReasoner, RSThinker} have made significant progress in grounding performance. Although reasoning modules are incorporated, the task formulation still focuses on aligning a single referring expression with one target region. In this sense, RS grounding is still treated as a matching problem and largely confined to the single-entity level.
\par On our proposed ME-RSRG dataset, we conduct comprehensive benchmarking with a series of representative foundation models to systematically evaluate existing grounding paradigms under the multi-entity reasoning setting. The main contributions of this paper are summarized as follows:
\begin{itemize}
\item We introduce ME-RSRG, a new and challenging benchmark dataset for multi-entity reasoning grounding in remote sensing. To the best of our knowledge, it is the first dataset that explicitly models multiple entities and their semantic roles, advancing research on reasoning grounding in remote sensing.
\item We propose an Entity-Aware Reasoning (EAR) framework based on visual-linguistic foundation models. It integrates supervised fine-tuning and entity-aware reward-driven GRPO. EAR provides benchmarking performance for our proposed ME-RSRG dataset.
\item Extensive experiments verify the effectiveness of EAR in structured relational grounding. With optimization under the EAR framework, the Qwen2.5-VL series achieves over 10\% improvement in mAcc@0.5 value, and all evaluated visual-linguistic foundation models show consistent gains.
\end{itemize}
\section{Related Work}
\subsection{Visual Grounding in Remote Sensing}
\par Visual grounding is first explored in natural scene images to localize objects described by language expressions. Representative methods~\cite{RefCOCO,RefCOCOg,Referitgame,Grounding-Dino,TransVG++} focus on cross-modal representation learning and region-text matching through notable architectures and fusion mechanism. Inspired by these advances in natural images, visual grounding has recently been extended to the remote sensing field. \cite{RSVG} is an early pioneering work that formally defines the remote sensing visual grounding task. To tackle this task, it proposes both a benchmark dataset and the GeoVG method. RSVG-HR~\cite{RSVG-HR} dataset is a redesigned language expression using high-resolution remote sensing images in the RSVG dataset~\cite{RSVG}. Zhan \textit{et al.}~\cite{DIOR-RSVG} construct the DIOR-RSVG benchmark dataset and propose a multigranularity visual-language fusion (MGVLF) method to address scale variation and cluttered background challenges in remote sensing scenes. Hu \textit{et al.}~\cite{Effi-Grounding-Dino} extend remote sensing visual grounding to the open-set setting. To address the limitations of single-object grounding, Efficient Grounding DINO is proposed to improve multi-object grounding in remote sensing. Liu \textit{et al.}~\cite{LGFormer} introduce a Language-Guided Transformer (LGFormer) with hybrid visual-textual representations for remote sensing visual grounding under complex scenes.
\subsection{Visual-Linguistic Reasoning Models}
\par Visual-linguistic reasoning models focus on integrating visual and textual information to support joint understanding on visual scenes. While SFT remains the dominant training paradigm for visual-language models, recent works explore reinforcement fine-tuning (RFT)~\cite{PPO, GRPO, DPO} as a complementary strategy to shape model reasoning through reward-driven optimization. Factually Augmented RLHF (Fact-RLHF)~\cite{Fact-RLHF} enhances the reward model with additional factual cues, including image captions and ground-truth multiple-choice options. RLHF-V~\cite{RLHF-V} addresses hallucination issues in multimodal large language models by aligning model behavior with fine-grained human feedback. It improves factual grounding and trustworthiness across multiple benchmarks. POVID~\cite{POVID} mitigates hallucinations by automating preference tuning, leveraging scalable feedback generation and optimization via Direct Preference Optimization (DPO). Recent advances show that Group Relative Policy Optimization (GRPO)~\cite{GRPO} has been widely explored in visual-linguistic reasoning tasks. Visual-RFT~\cite{Visual-RFT} extends reinforcement fine-tuning with verifiable rewards to visual tasks, showing stronger generalization than SFT across multiple vision and reasoning benchmarks. So-Fake-R1~\cite{So-Fake} explores ``Cold Start-GRPO'' strategy to encourage interpretable visual rationales under realistic distribution shifts. Vision-R1~\cite{Vision-R1} also utilizes reinforcement learning for multimodal reasoning. It further proposes Progressive Thinking Suppression Training (PTST) to stabilize multimodal reasoning learning and mitigate overthinking during reinforcement fine-tuning.
\subsection{Visual-Linguistic Reasoning in Remote Sensing}
Visual-linguistic reasoning has mainly been explored in natural scenes. Applying it to remote sensing remains challenging due to large-scale layouts and complex spatial relations. SkySense-O~\cite{SkySense-O} advances open-world RS image understanding by introducing the large-scale ``Sky-SA'' fine-grained pixel-level dataset and a vision-centric visual-linguistic modeling paradigm. SegEarth-R1~\cite{SegEarth-R1} is a first-proposed language-guided segmentation framework that advances geospatial pixel reasoning by enabling implicit query-driven spatial understanding. RemoteReasoner~\cite{RemoteReasoner} is a unified RS reasoning framework that enables a single model to handle diverse remote sensing reasoning tasks from natural language queries without task-specific fine-tuning. RSThinker~\cite{RSThinker} investigates structured and verifiable multi-step reasoning for remote sensing analysis via large-scale rationale supervision and reinforcement learning.
\begin{figure}[!t]
  \centering
\includegraphics[width=1.0\textwidth]{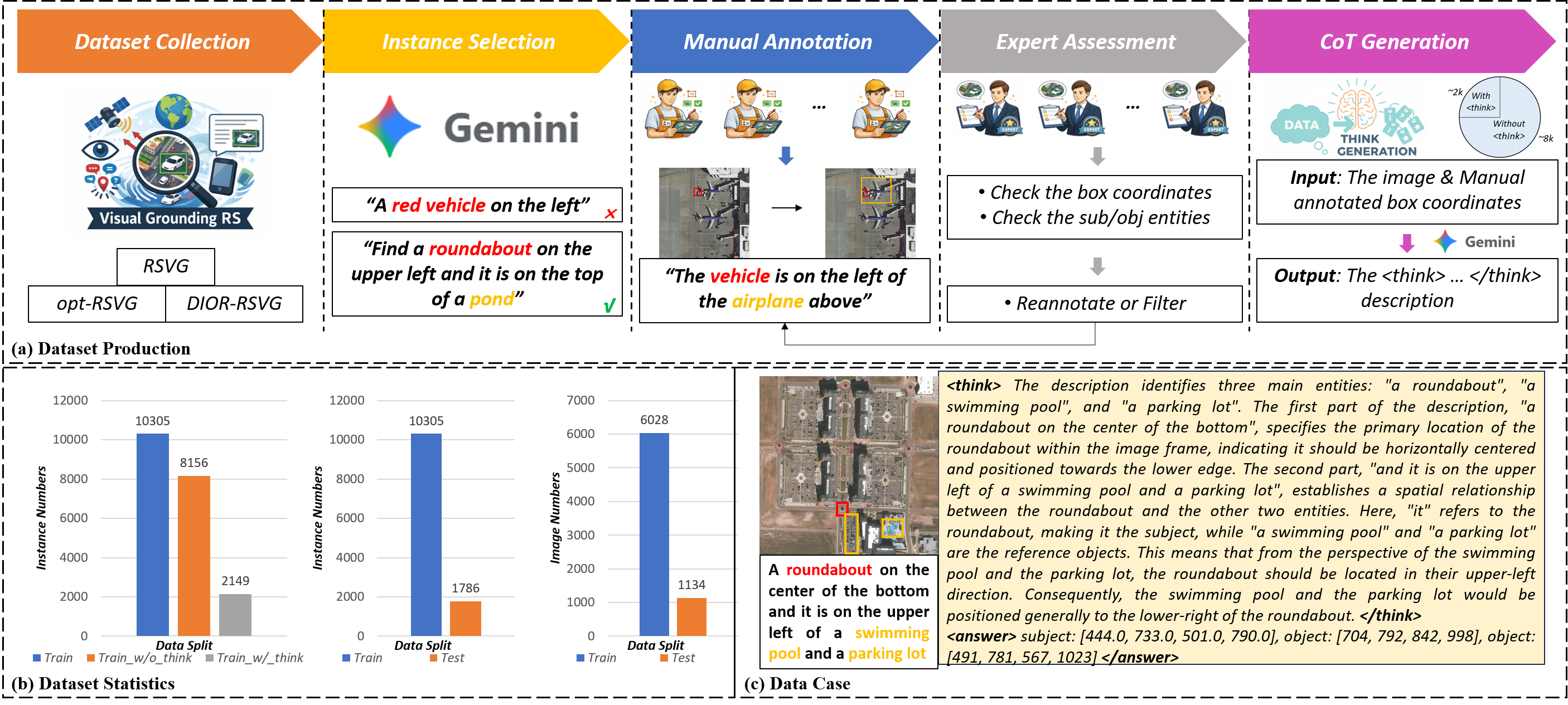}
  \caption{\textbf{Overview of ME-RSRG dataset.} It includes (a) Dataset production, (b) Dataset statistics, and (c) Data case.}
  \label{Fig2}
\end{figure}
\section{Proposed Dataset: ME-RSRG}
\subsection{Dataset Production}
As shown in Fig.~\ref{Fig2} (a), the production of ME-RSRG follows a 5-step pipeline to ensure data quality. It includes dataset collection, instance selection, manual annotation, expert assessment, and CoT generation. 
\par \noindent \textbf{Dataset Collection.} We collect data from three benchmark RS visual grounding datasets: RSVG-HR~\cite{RSVG-HR}, DIOR-RSVG~\cite{DIOR-RSVG}, and OPT-RSVG~\cite{opt-RSVG}. RSVG-HR is built on the very high-resolution images of RSVG~\cite{RSVG}, with rewritten language annotations using a clearer protocol of absolute position and relative position. It contains 2,650 image-language pairs across 7 categories, with the image size as 1024$\times$1024. DIOR-RSVG is built upon the DIOR dataset~\cite{DIOR} and covers 20 object categories. It contains 38,320 language expressions over 17,402 images of size 800$\times$800. OPT-RSVG consists of 25,452 RS images and 48,952 image-query pairs among 14 object categories. Since part of the OPT-RSVG dataset is derived from DIOR, we prevent data leakage during the collection process. 
\par \noindent \textbf{Instance Selection.} In this step, we perform expression-level analysis to select data containing explicit subject-object structures and multiple entities. As shown in Fig.~\ref{Fig2} (a), we employ Gemini-2.5-Pro~\cite{Gemini} to filter the collected expressions. Only those describing relational dependencies between at least two entities are retained. In addition, the detailed prompts used for Gemini in the data selection process will be provided in the supplementary material.
\par \noindent \textbf{Manual Annotation.} After instance selection, we conduct manual annotation to construct multi-entity grounding instances. We inherit the original bounding box annotations and language descriptions from RSVG-HR, DIOR-RSVG, and OPT-RSVG. Based on these single-subject annotations, annotators further identify and label corresponding object entities within the same image. At the current research stage, our annotation protocol focuses on ``single-subject single-object'' and ``single-subject multiple-objects'' settings. This step also includes filtering and correcting inappropriate expressions identified in the previous stage to ensure semantic consistency and annotation quality.
\par \noindent \textbf{Expert Assessment.} As depicted in Fig.~\ref{Fig2} (a), domain experts (trained students and annotators) conduct a strict quality review of the annotations. Each instance undergoes sequential review by two experts. They check the accuracy of bounding box coordinates and verify the correctness of subject and object entity assignments. Samples that fail to meet the quality criteria are either re-annotated (back to ``Manual Annotation'' step) or removed from the dataset.
\par \noindent \textbf{CoT Generation.} In the final step, we generate reasoning traces for a subset of the data ($\sim$20\% of training set). Given the image and manually annotated bounding boxes, we employ Gemini-2.5-Pro~\cite{Gemini} to produce reasoning descriptions in the tag of $<think>...</think>$. These descriptions provide intermediate reasoning signals to support reasoning-oriented training. The exact prompt used for CoT generation with Gemini will be detailed in the supplementary material.
\subsection{Dataset Analysis}
\par \noindent \textbf{Dataset Statistical Analysis.} As depicted in Fig.~\ref{Fig2} (b), ME-RSRG contains 7,162 images and 12,091 image-text instances in total. A single image may correspond to multiple descriptions, reflecting diverse multi-entity relations within the same scene. The dataset is split into 10,305 training instances and 1,786 test instances, covering 6,028 and 1,134 images, respectively. Among the 10,305 training instances, 2,149 are annotated with explicit $<think>...</think>$ reasoning tags, while the remaining 8,156 provide only grounding annotations without intermediate reasoning traces. The CoT-annotated samples are typically used for SFT as a cold-start initialization.
\par \noindent \textbf{Dataset Case Analysis.} As depicted in Fig.~\ref{Fig2} (c), we analyze the data characters from the following aspects. First, large-scale spatial layouts are prominent. Object scales vary greatly within a scene. Their wide spatial distribution makes spatial configuration a key cue for grounding. The identity of a subject often depends on its relative position, or functional association with other entities. This demands explicit modeling of subject-object roles. Second, multi-entity ambiguity increases the difficulty. A single image may contain many similar instances, such as multiple ``roundabouts'' shown in Fig.~\ref{Fig2} (c). Accurate grounding relies on relational disambiguation rather than isolated appearance cues. Third, the dataset supports step-by-step subject-object reasoning with partial CoT supervision. It also requires simultaneous localization of object entities, ensuring role-consistent and interpretable multi-entity grounding.
\begin{table*}[!t]
  \scriptsize
  \centering
  \caption{\textbf{Comparison between ME-RSRG and other related datasets.} In the table, ``Im.'', and ``Ins.'' respectively denote the image and instance (image-text pair).}
  \scalebox{0.98}{
  \begin{tabular}{p{0.22\textwidth}|p{0.17\textwidth}|p{0.08\textwidth} |p{0.16\textwidth}|p{0.20\textwidth}|p{0.13\textwidth}}
    \hline\hline
    \makecell[c]{\textbf{Dataset}} & \makecell[c]{\textbf{Im./Ins. Num.}} & 
    \makecell[c]{\textbf{Entity}} & \makecell[c]{\textbf{Im. Size}} & \makecell[c]{\textbf{Sources}} & \makecell[c]{\textbf{Tasks}}\\
  \hline \hline
    \rowcolor{gray!30!} \multicolumn{6}{c} {\textbf{Remote Sensing Visual Grounding Datasets}}  \\
    \hline
    \makecell[c]{RSVG~\cite{RSVG}} & \makecell[c]{4,239/7,933} & \makecell[c]{Single} & \makecell[c]{$1024\times1024$} & \makecell[c]{Google Earth} & \makecell[c]{VG} \\
    \hline
    \makecell[c]{RSVG-HR~\cite{RSVG-HR}} & \makecell[c]{4,239/2,650} & \makecell[c]{Single} & \makecell[c]{$1024\times1024$} & \makecell[c]{Google Earth} & \makecell[c]{VG} \\
    \hline
    \makecell[c]{DIOR-RSVG~\cite{DIOR-RSVG}} & \makecell[c]{17,402/38,320} & \makecell[c]{Single} & \makecell[c]{$800\times800$} & \makecell[c]{DIOR~\cite{DIOR}} & \makecell[c]{VG} \\
    \hline
    \makecell[c]{OPT-RSVG~\cite{opt-RSVG}} & \makecell[c]{25,452/48,952} & \makecell[c]{Single} & \makecell[c]{$\sim$152$^2$-$\sim$10569$^2$} & \makecell[c]{DIOR~\cite{DIOR}, \\ SPCD~\cite{SPCD}, \\ HRRSD~\cite{HRRSD}} & \makecell[c]{VG} \\
    \hline
    \makecell[c]{VRSBench-Ref~\cite{VRSBench}} & \makecell[c]{29,614/52,472} & \makecell[c]{Single} & \makecell[c]{$512\times512$} & \makecell[c]{DIOR~\cite{DIOR}, \\ DOTA-v2~\cite{DOTA-v2}} & \makecell[c]{VG} \\
    \hline
  \rowcolor{gray!30!} \multicolumn{6}{c} {\textbf{Remote Sensing Reasoning Datasets}}  \\
    \hline
    \makecell[c]{RemoteReasoner~\cite{RemoteReasoner}} & \makecell[c]{5,434/-} & \makecell[c]{Single} & \makecell[c]{123$^2$-7617$^2$} & \makecell[c]{EarthReason~\cite{SegEarth-R1}} & \makecell[c]{\textbf{RG}, VQA, \\ Captioning} \\
    \hline
    \makecell[c]{Geo-CoT380k~\cite{RSThinker}} & \makecell[c]{-/70,711} & \makecell[c]{Single} & \makecell[c]{$512\times512$, \\ $800\times800$} & \makecell[c]{DIOR-RSVG~\cite{SegEarth-R1}, \\ VRSBench-Ref~\cite{VRSBench}} & \makecell[c]{\textbf{RG}, VQA, \\ Captioning, \\ Counting, \\
    Cls, Det} \\
    \hline
    \makecell[c]{ME-RSRG (ours)} & \makecell[c]{7,162/12,091} & \makecell[c]{\textbf{Multi.}} & \makecell[c]{$1024\times1024$, \\ $800\times800$, \\ $\sim$152$^2$-$\sim$10569$^2$} & \makecell[c]{DIOR-RSVG~\cite{SegEarth-R1}, \\ RSVG-HR~\cite{RSVG-HR}, \\ OPT-RSVG~\cite{opt-RSVG}} & \makecell[c]{\textbf{RG}} \\
    \hline\hline  
\end{tabular}
}
  \label{Tab1}
\end{table*}
\par \noindent \textbf{Dataset Comparison.} Tab.~\ref{Tab1} presents a comprehensive comparison between ME-RSRG and existing remote sensing visual grounding and reasoning datasets. Most remote sensing visual grounding datasets are constructed under a single-entity paradigm. The VG tasks built on these datasets mainly focus on perception-level alignment between textual queries and individual objects. Recent reasoning-oriented datasets, such as RemoteReasoner and Geo-CoT380k, extend the task scope to reasoning grounding and other reasoning tasks. However, they still rely on single-entity annotations and do not explicitly model multiple entities with structured subject-object roles.
\par In contrast, our proposed ME-RSRG is explicitly designed for multi-entity reasoning grounding. Each instance contains a subject entity and one or more object entities with structured relational descriptions. It further provides explicit ``think'' annotations for partial training data. The dataset emphasizes challenges from both scale variation and domain diversity, covering wide-ranging object sizes as well as multiple image sources and resolutions. To the best of our knowledge, it is the first remote sensing dataset that explicitly models multiple entities and their semantic roles within structured relational descriptions.
\begin{figure}[!t]
  \centering
\includegraphics[width=1.0\textwidth]{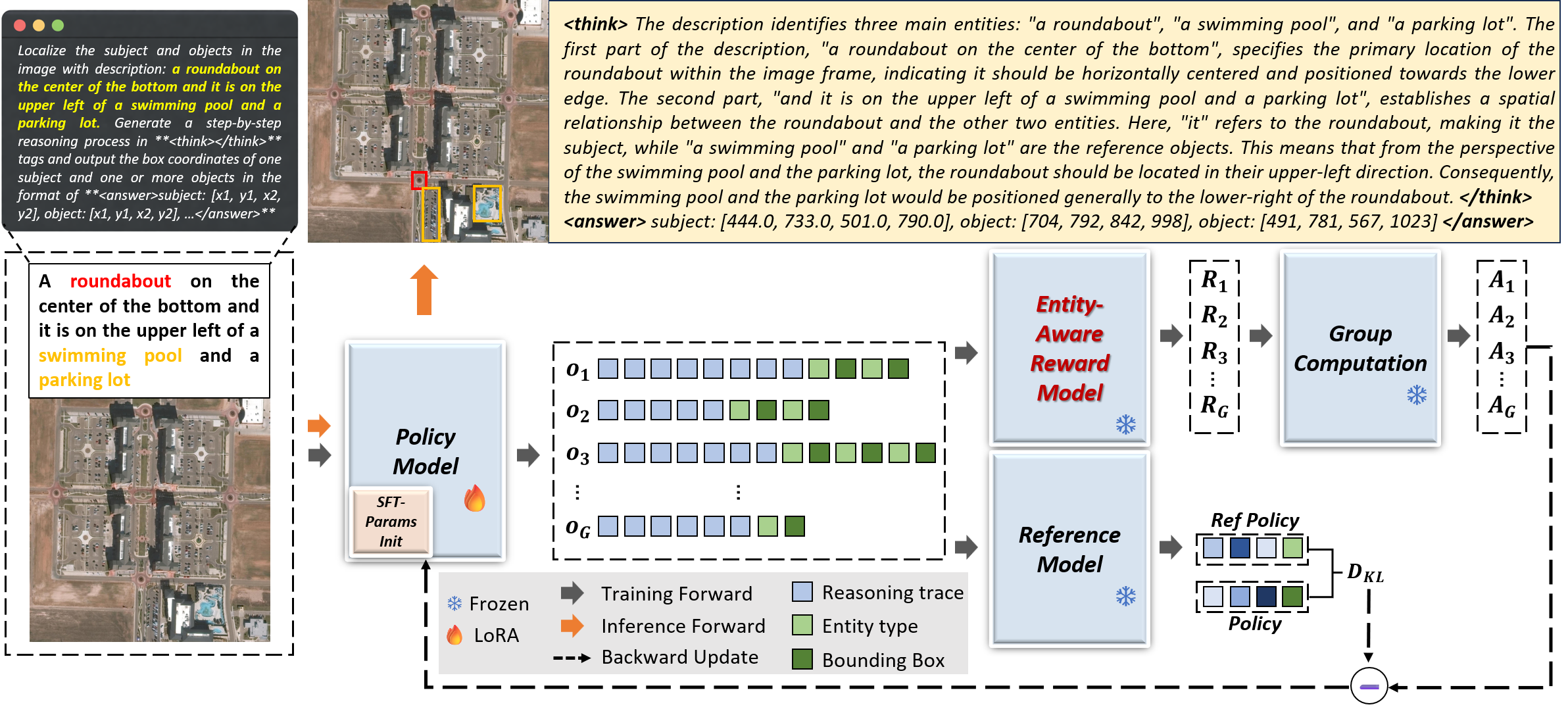}
  \caption{\textbf{Overview of EAR framework.} We adopt a two-stage optimization strategy: SFT initialization and entity-aware reward-driven GRPO refinement.}
  \label{Fig3}
\end{figure}
\section{Methodology}
\par To support multi-entity reasoning grounding on ME-RSRG, we propose an Entity-Aware Reasoning (EAR) framework as a dedicated benchmarking framework. EAR integrates visual-linguistic encoding with structured reasoning generation, explicitly modeling subject–object roles and their relations.
\subsection{Overall Architecture}
As illustrated in Fig.~\ref{Fig3}, the EAR framework is built on two core models: a policy model and a reference model. The policy model serves as the main reasoning engine. It is initialized from a visual-linguistic foundation model (e.g., Qwen-VL series~\cite{Qwen2.5-VL, Qwen3-VL}) and takes the image and referring expression as input. Given each input, the policy model samples multiple structured completions ($\{o_{i}\}_{i=1}^{G}$), including the reasoning trace (``think'') and the ``answer'' representing entity type and bounding box. These sampled completions are then evaluated by our proposed entity-aware reward model, which assigns rewards ($\{R_{i}\}_{i=1}^{G}$) based on grounding accuracy for different entity roles and their relational consistency. These rewards are further normalized within each sample group to compute relative advantages ($\{A_{i}\}_{i=1}^{G}$). In parallel, a frozen copy of the initial policy is maintained as the reference model. The KL divergence ($D_{KL}$) between the current policy and this reference distribution is computed to regularize updates and stabilize training.
\subsection{Entity-Aware Reward-Driven GRPO}
Based on the overall architecture, we propose the core component of EAR: an Entity-Aware Reward model. Specifically, it consists of three components: (1) \textbf{Two-level format reward.} It ensures structured reasoning and answer outputs. (2) \textbf{An entity-aware grounding accuracy reward.} This reward evaluates subject and object localization separately. (3) \textbf{A relational consistency reward.} It assesses the coherence of predicted inter-entity relations.
\par \noindent \textbf{Two-Level Format Reward.} To guarantee stable structured generation under the reasoning grounding setting, we design a two-level format reward ($R_{fmt}$). It constrains both the global output structure and the entity-level answer format.
\par The first level (\textit{structural tag reward}) enforces the overall output template. Each completion must strictly follow the form as $<think>...</think> <answer>...</answer>$. The $<think>$ tag contains the reasoning trace, and the $<answer>$ block contains the final grounding results.
\par The second level (\textit{entity-aware answer reward}) verifies whether the $<answer>$ block follows the predefined entity representation form: ``$subject: [(x1, y1), (x2, y2)]$,
$object: [(x1, y1), (x2, y2)], ...$''. Each entity is assigned a role label (e.g., subject or object) and a bounding box defined by two coordinate points. This reward is granted only when at least one correctly formatted entity is detected.
\par The final format reward is computed as the sum of the two levels (Eq.~\ref{Eq1}). 
\begin{equation}
\begin{split}
    R_{fmt} &= \lambda_{1} * \mathbbm{1}( \text{structural tag format satisfied}) \\ &+ \lambda_{2} * \mathbbm{1}(\text{valid entity format detected})
\end{split}
\label{Eq1}
\end{equation}
where $\mathbbm{1}(\cdot)$ denotes the indicator function. We set $\lambda_{1} = 0.3$ and $\lambda_{2} = 0.3$, yielding a maximum format reward of 0.6 when both conditions are satisfied. If only one condition holds, the reward is 0.3; otherwise, it is 0. This partial-credit scheme facilitates progressive learning of structural and entity-level formatting. By jointly constraining reasoning structure and entity output pattern, the two-level format reward provides a reliable foundation for subsequent entity-aware grounding accuracy and relational consistency reward within the GRPO stage.
\par \noindent \textbf{Entity-Aware Grounding Accuracy Reward.} To explicitly evaluate grounding quality under the multi-entity setting, we design an entity-aware grounding accuracy reward ($R_{ent}$). It operates on the prediction of the entities extracted from the validated block $<answer>$ (shown in Eq.~\ref{Eq2}).
\begin{equation}
    R_{ent}
    = \frac{1}{N}\sum_{i=1}^{N}\alpha_{i}r(iou_{i}),
   \quad
    r(iou_{i}) = 
    \begin{cases}{}
        1.0, iou_{i} > 0.75, \\ 
        0.8, iou_{i} > 0.5, \\
        0.4, iou_{i} > 0.25, \\
        0, otherwise
    \end{cases}
\label{Eq2}
\end{equation}
where $N$ denotes the total number of predicted entities (including the subject and all objects). $\alpha_{i}$ is the role-specific weight for the $i^{th}$ entity (1.5 for the subject and 1.25 for objects). $iou_{i}$ represents the intersection-over-union between the predicted and ground-truth bounding boxes of the $i^{th}$ entity. The function $r(\cdot)$ maps the IoU values to discrete base scores according to predefined thresholds.
\par This reward function is designed with the following principles. (1) The role-specific weighting encourages subject matching. In multi-entity referring expressions, the subject serves as the primary grounding target, while object entities provide relational constraints. (2) The reward assigns partial credit once IoU exceeds 0.25 and increases progressively at higher thresholds. This dense feedback signal encourages incremental improvement. (3) We adopt discrete IoU thresholds to promote precise localization. Compared with continuous scores, threshold-based rewards ease credit assignment in multi-entity settings by enforcing clear correctness boundaries over minor geometric noise.
\par \noindent \textbf{Relational Consistency Reward (Bonus Reward).} Beyond entity-aware grounding accuracy, we introduce a relational consistency reward ($R_{rel}$) as an instance-level bonus to explicitly reinforce relation-level grounding completeness.
\par We denote the set of subject entities as $\mathcal{S}$ and the set of object entities as $\mathcal{O}$. In our current research setting, $|\mathcal{S}| = 1$, while $|\mathcal{O}| \ge 1$. For each ground-truth entity $k \in \mathcal{S} \cup \mathcal{O}$, we define a binary matching indicator ($m_{k}$) to measure the predicted box and the corresponding ground-truth box of $k^{th}$ entity. The relational consistency reward is formulated in Eq.~\ref{Eq3}.
\begin{equation}
    \begin{split}
        R_{rel} &= \beta_{1}\mathbbm{1}(\sum_{i \in \mathcal{S}} m_i \ge 1 \;\land\; \sum_{j \in \mathcal{O}} m_j \ge 1) \\ &+ \beta_{2}\mathbbm{1}(\sum_{j \in \mathcal{O}} m_j \ge 2),
    \end{split}
    \quad
    m_{k} = 
    \begin{cases}{}
        1, iou_{k} > 0.25, \\ 
        0, otherwise
    \end{cases}
\label{Eq3}
\end{equation}
where $\beta_1$ and $\beta_2$ are bonus values (set to 0.3 in our implementation), $\mathbbm{1}(\cdot)$ denotes the indicator function that outputs 1 when the condition holds and 0 otherwise.
\par As shown in Eq.~\ref{Eq3}, the first term encourages relational completeness by requiring both the subject and at least one object to be correctly matched. This ensures that grounding captures a valid subject-object relation rather than isolated entity localization. The second term promotes relational grounding by rewarding correct matching of multiple object entities. It encourages the model to capture richer relational structures when multiple objects are involved.
\par As a whole, the total reward ($R_{total}$) is calculated in Eq.~\ref{Eq4}.
\begin{equation}
    R_{total} = R_{fmt} + R_{ent} + R_{rel}
\label{Eq4}
\end{equation}
\subsection{Optimization}
As shown in Fig.~\ref{Fig3}, EAR adopts a two-stage optimization strategy. In the first stage, SFT serves as a cold-start initialization. In the second stage, entity-aware reward-driven GRPO is employed to further refine the policy model. Preliminary knowledge for two-stage optimization is provided in the supplementary material. 
\par \noindent \textbf{Stage I: Supervised Fine-Tuning (SFT).} In the first stage, we perform SFT to initialize the policy model with structured reasoning capability. Given an image-expression pair, the model is trained to generate a unified output sequence that includes both the reasoning trace ($<think>$ tag) and the final subject-object grounding result ($<answer>$ tag). This stage teaches the model the required output structure, entity identification, and coarse grounding behavior.
\par \noindent \textbf{Stage II: Entity-Aware Reward-Driven GRPO.} Building upon the SFT initialization, we employ GRPO to transition the model from basic pattern matching to structured reasoning. For each input, the policy samples a group of independent completions. These outputs are evaluated by our proposed Entity-Aware Reward function, which integrates format constraints, grounding accuracy, and relational consistency bonuses. Specifically, this reward mechanism penalizes role confusion by validating the correspondence between entities and their spatial coordinates. By normalizing each $R_{i}$ against the group mean, the model derives the advantage $A_{i}$ through group-relative comparison. This mechanism prioritizes reasoning paths with higher entity localization precision. Overall, this stage mitigates ``path drift'' in complex multi-entity scenarios, ensuring the model maintains both structural integrity and high grounding performance.
\section{Experimental Results}
\subsection{Experimental Setup}
\par \noindent \textbf{Dataset.} We conduct extensive benchmarking using our proposed ME-RSRG dataset. For the initial cold-start stage, a subset of 2,149 instances containing annotated reasoning chains (labeled with <think> tags) is utilized for $1^{st}$ stage SFT. In the subsequent reinforcement learning stage, we expand the training to 10,305 instances for GRPO, which encompasses the entire training set. Notably, in the GRPO process, all reasoning CoT annotations are removed. 
\par \noindent \textbf{Implementation Details.} To establish a comprehensive benchmark on our proposed ME-RSRG dataset, we mainly select the notable Qwen-VL series~\cite{Qwen2-VL, Qwen2.5-VL, Qwen3-VL} and InternVL series~\cite{InternVL-2.5, InternVL-3.5} as the primary policy models. All experiments are implemented using the ms-swift framework~\cite{ms-swift}, executed on single NVIDIA A800 GPU (80GB). More details are provided in the supplementary material.
\par \noindent \textbf{Evaluation Metrics.} To evaluate the performance of EAR framework and provide comprehensive benchmarking baseline on ME-RSRG dataset, we employ mean accuracy at an IoU threshold of 0.5 (mAcc@0.5) as the primary metric. We further report Acc@0.5\_sub and Acc@0.5\_obj, which respectively measure the grounding precision of primary entity (subject) and target landmarks (objects).
\subsection{Main Experimental Results}
\par \noindent \textbf{Baseline Capability.} As shown in Tab.~\ref{Tab2}, we benchmark the zero-shot performance of different visual-linguistic foundation models. Overall, most models exhibit extremely limited multi-entity grounding ability without task-specific fine-tuning. Even large-scale models provide only modest gains. For example, Qwen3-VL-8B achieves the strongest baseline performance with an mAcc@0.5 of 21.72\%, yet the overall accuracy remains limited. This highlights the challenges of the ME-RSRG dataset, which requires precise role identification, relational reasoning, and accurate spatial localization under multi-entity settings.
\par \noindent \textbf{Impact of SFT Initialization.} In Tab.~\ref{Tab2}, we find that all foundation models show substantial improvements over their baseline counterparts after SFT. For instance, Qwen2.5-VL-7B improves by 22.99\% in mAcc@0.5, and InternVL3.5-4B rises sharply from 0.51\% to 32.80\%. These consistent gains demonstrate that SFT effectively equips the models with structured output formatting, basic entity role awareness, and coarse grounding capability.
\par \noindent \textbf{Impact of Optimization within EAR framework.} As reported in Tab.~\ref{Tab2}, Building upon SFT initialization, the full EAR optimization consistently brings further performance gains across most models. The Qwen2.5-VL series shows particularly significant improvements, each exceeding $\sim$10\% in mAcc@0.5 after EAR optimization. All models exhibit performance improvements compared with their SFT counterparts. Compared with SFT alone, EAR yields balanced gains on both subject and object accuracy. It indicates that reward-driven optimization effectively enhances entity-aware grounding and relational consistency.
\begin{table*}[!t]	
	\centering
	\caption{\textbf{Performance of representative methods on the ME-RSRG dataset.} Results for ``sub.'', ``obj.'', and ``mean'' denote Acc@0.5\_sub (\%), Acc@0.5\_obj (\%), and mAcc@0.5 (\%), respectively. The best-performing results are highlighted in bold.}
    \scalebox{1.0}{
    \begin{tabular}{c c c c c c c c c c}
    \hline \hline
  \makecell[c]{\multirow{2}{*}{\shortstack{Methods \\ (policy models)}}} &  \multicolumn{3}{c}{\makecell[c]{Baseline}} & \multicolumn{3}{c}{\makecell[c]{SFT}} & \multicolumn{3}{c}{\makecell[c]{EAR}}
  \\ \cmidrule(r){2-10}
  {} & \makecell[c]{sub.} & \makecell[c]{obj.} & \makecell[c]{mean} & \makecell[c]{sub.} & \makecell[c]{obj.} & \makecell[c]{mean} & \makecell[c]{sub.} & \makecell[c]{obj.} & \makecell[c]{mean}
  \\ \hline
   \rowcolor{gray!30!} \multicolumn{10}{c} {\textbf{Qwen-VL series}}  
   \\ \hline
   \makecell[c]{Qwen2-VL-2B~\cite{Qwen2-VL}} & 1.40 & 1.15 & 1.27 & 23.18 & 23.08 & 23.13 & 27.55 & 22.93 & 25.16 \\ \hline
   \makecell[c]{Qwen2.5-VL-3B~\cite{Qwen2.5-VL}} & 7.45 & 10.81 & 9.18 & 27.04 & 30.43 & 28.79 & 38.63 & 42.08 & 40.41 \\ \hline
   \makecell[c]{Qwen2.5-VL-7B~\cite{Qwen2.5-VL}} & 11.31 & 9.44 & 10.35 & 31.75 & 34.84 & 33.34 & \textbf{44.40} & \textbf{48.90} & \textbf{46.72} \\ \hline
   \makecell[c]{Qwen3-VL-4B~\cite{Qwen3-VL}} & 20.95 & 13.80 & 17.25 & \textbf{36.39} & \textbf{35.89} & \textbf{36.13} & 38.52 & 36.83 & 37.65 \\ \hline
   \makecell[c]{Qwen3-VL-8B~\cite{Qwen3-VL}} & \textbf{26.82} & \textbf{16.95} & \textbf{21.72} & 33.76 & 34.89 & 34.34 & 37.96 & 36.94 & 37.43 \\ \hline
   \rowcolor{gray!30!} \multicolumn{10}{c} {\textbf{InternVL series}} \\ \hline
   \makecell[c]{InternVL2.5-4B~\cite{InternVL-2.5}} & 1.68 & 1.57 & 1.63 & 6.94 & 5.55 & 6.18 & 9.18 & 9.97 & 9.59 \\ \hline
   \makecell[c]{InternVL3.5-4B~\cite{InternVL-3.5}} & 0.50 & 0.52 & 0.51 & 34.49 & 31.22 & 32.80 & 38.02 & 35.31 & 36.62 \\ \hline
   \makecell[c]{InternVL3.5-8B~\cite{InternVL-3.5}} & 2.35 & 2.31 & 2.33 & 35.33 & 32.32 & 33.78 & 39.08 & 34.89 & 36.92 \\ \hline
   \rowcolor{gray!30!} \multicolumn{10}{c} {\textbf{Other Visual-Linguistic Foundation Models}}  \\ \hline
   \makecell[c]{LLaVA-1.5-7b~\cite{llava}} & 1.29 & 1.21 & 1.25 & 3.42 & 3.25 & 3.33 & 7.95 & 7.97 & 7.96 \\ \hline
   \makecell[c]{LLaVA-OneVision~\cite{LLaVA-OneVision}} & 0 & 0.05 & 0.03 & 9.80 & 6.24 & 7.96 & 18.14 & 14.59 & 16.31 \\
  \hline \hline
   \end{tabular}
 }
 \label{Tab2}
 \end{table*}
\subsection{Ablation Study}
\par \noindent \textbf{Effectiveness of Each Optimization Mechanism.} From the results in Tab.~\ref{Tab3}, the ``GRPO-only'' mode does not yield satisfactory performance. Removing SFT and directly applying GRPO leads to clear degradation for both models. For instance, Qwen2.5-VL-3B drops from 28.79 to 15.49 in mAcc@0.5. It shows that reinforcement learning without structured initialization cannot establish stable role-aware grounding. This verifies that SFT is a necessary prerequisite for effective reward-driven optimization. Notably, Qwen3-VL-4B is more robust to optimization variations. Although ``GRPO-only'' still underperforms SFT, the gap is smaller than that of Qwen2.5-VL-3B. In both models, combining SFT and GRPO yields the best results, confirming the effectiveness of the two-stage optimization mechanism with EAR framework.
\par \noindent \textbf{Effectiveness of Each Reward Function.} From the results in Tab.~\ref{Tab4}, entity-aware grounding accuracy and the second-level format reward are included in all settings, as they are responsible for entity extraction. Without them, the task objective is ill-defined. Removing the first-level format reward leads to consistent performance drops. This reward regularizes the structured reasoning process. Without it, the $<think>$ trace tends to collapse in later stages, and training becomes unstable due to reward fluctuations. Removing the relational consistency reward causes the model to predict only a single object and overlook relational completeness. It weakens both subject-object and object-object coordination. Consequently, both subject and object accuracies decline, which proves the importance of maintaining multi-entity consistency in remote sensing reasoning grounding.
\begin{table}[t]
    \centering
    \begin{minipage}{0.48\textwidth}
        \centering
        \caption{Ablation study for optimization mechanism. Here, ``GRPO'' indicates the entity-aware reward-driven GRPO.}
        \label{Tab3}
        \scalebox{0.85}{
        \begin{tabular}{c|cc|ccc}
            \hline \hline
            method & SFT & GRPO & sub. & obj. & mean \\ \hline
            Qwen2.5-VL-3B & \checkmark & \ding{56} & 27.04 & 30.43 & 28.79 \\ 
            Qwen2.5-VL-3B & \ding{56} & \checkmark & 16.74 & 14.30 & 15.49 \\ 
            Qwen2.5-VL-3B & \checkmark & \checkmark & 38.63 & 42.08 & 40.41 \\ \hline
            Qwen3-VL-4B & \checkmark & \ding{56} & 36.39 & 35.89 & 36.13 \\
            Qwen3-VL-4B & \ding{56} & \checkmark & 32.19 & 31.64 & 31.91 \\
            Qwen3-VL-4B & \checkmark & \checkmark & 38.52 & 36.83 & 37.65 \\
            \hline \hline
        \end{tabular}
        }
    \end{minipage}
    \hfill 
    \begin{minipage}{0.48\textwidth}
        \centering
        \caption{Ablation study for reward functions. Entity-aware accuracy and second-level format rewards are used by default.}
        \label{Tab4}
        \scalebox{0.85}{
        \begin{tabular}{c|cc|ccc}
            \hline \hline
            method & $1^{st} R_{fmt}$ & $R_{rel}$ & sub. & obj. & mean \\ \hline
            Qwen2.5-VL-3B & \checkmark & \ding{56} & 33.48 & 33.21 & 33.34 \\ 
            Qwen2.5-VL-3B & \ding{56} & \checkmark & 34.32 & 39.14 & 36.81 \\ 
            Qwen2.5-VL-3B & \checkmark & \checkmark & 38.63 & 42.08 & 40.41 \\ \hline
            Qwen3-VL-4B & \checkmark & \ding{56} & 34.83 & 31.22 & 32.96 \\
            Qwen3-VL-4B & \ding{56} & \checkmark & 35.66 & 34.63 & 35.13 \\
            Qwen3-VL-4B & \checkmark & \checkmark & 38.52 & 36.83 & 37.65 \\
            \hline \hline
        \end{tabular}
        }
    \end{minipage}
\end{table}
\subsection{Visualization and Analysis}
\par
\begin{figure}[!t]
  \centering
\includegraphics[width=1.0\textwidth]{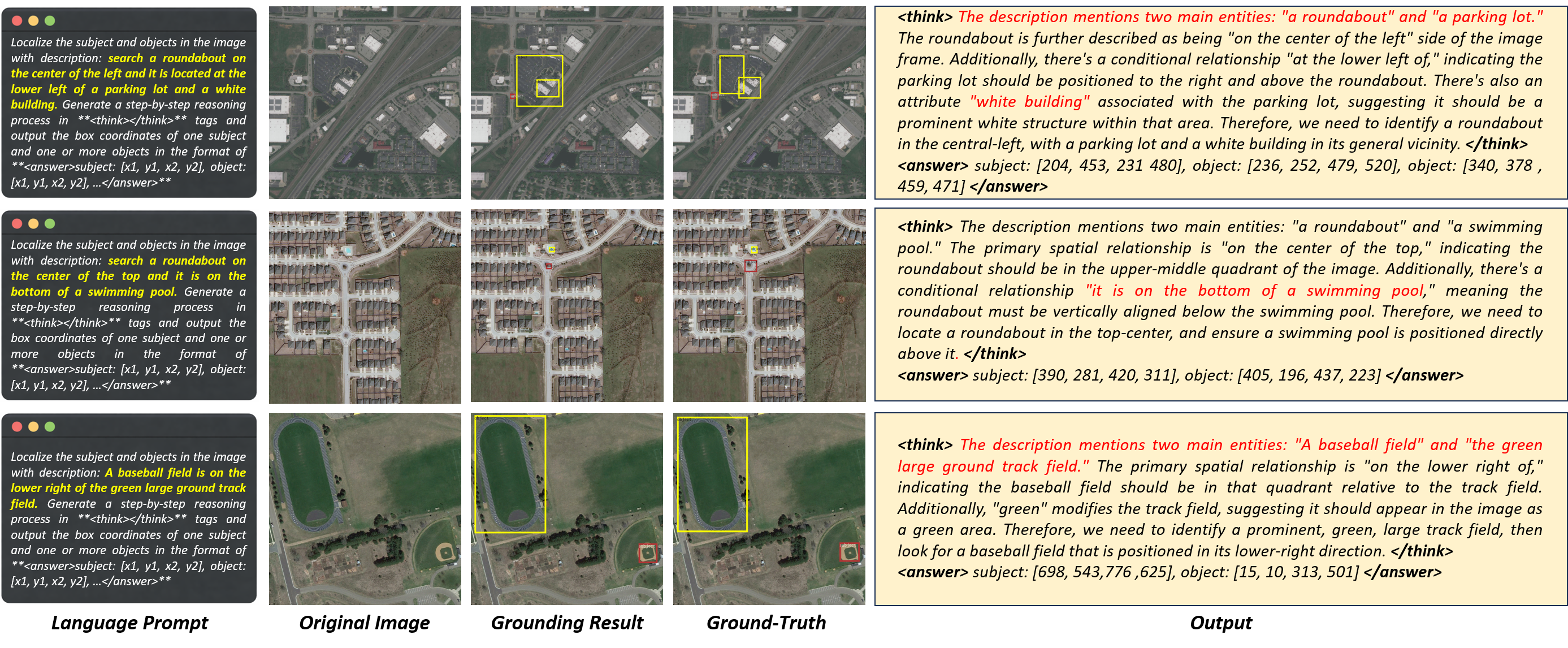}
  \caption{\textbf{Qualitative Results Visualization.}}
  \label{Fig4}
\end{figure}

 \noindent \textbf{Visualization on  Qualitative Results and Failure Cases.} We select the best-performed Qwen2.5-VL-7B for experiments. As shown in Fig.~\ref{Fig4}, the model shows strong ability across diverse scenarios. From top to bottom, the three cases demonstrate complementary strengths of the model. In the first case, the model accurately localizes multiple entities while preserving their spatial relationships. In the second case, it successfully identifies small targets under interference from similar structures. In the third case, the model can find entities with large size variation within the same scene, achieving precise multi-scale localization ability. 
\par
\begin{figure}[!t]
  \centering
  \begin{minipage}[t]{0.49\textwidth}
    \centering
    \includegraphics[width=\textwidth]{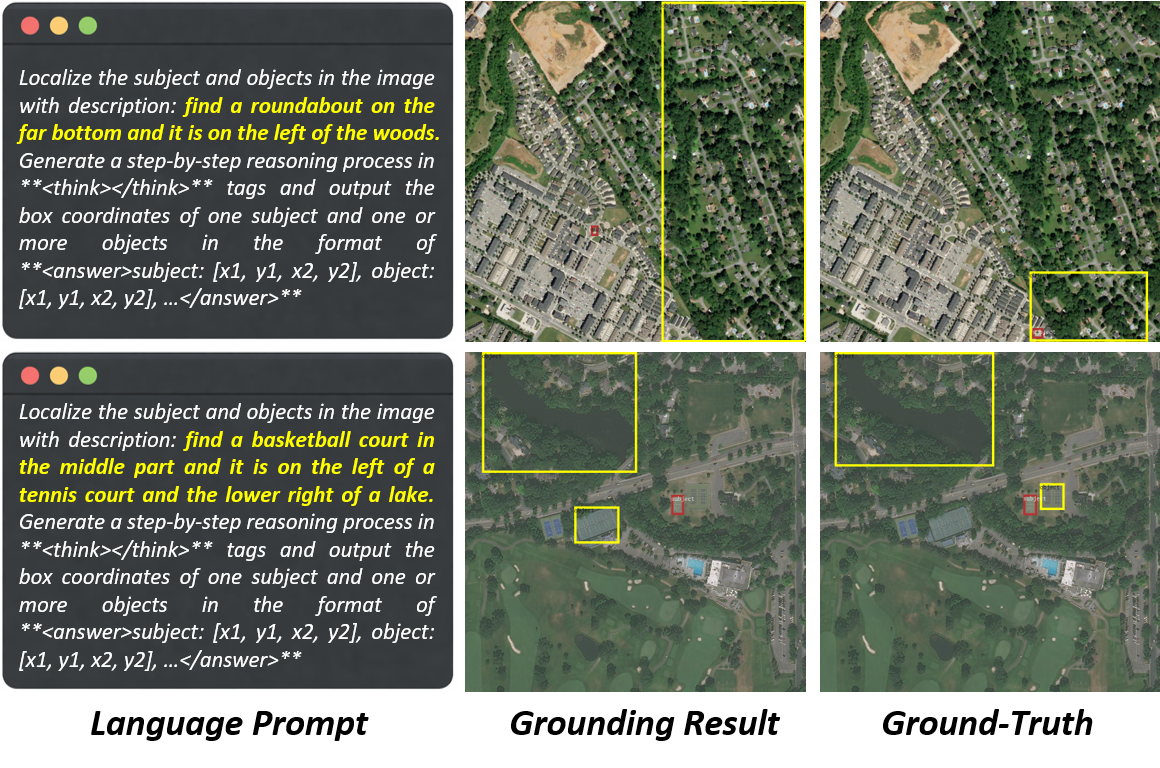}
    \caption{\textbf{Failure Cases Visualization.}}
    \label{Fig5}
  \end{minipage}
  \hfill
  \begin{minipage}[t]{0.49\textwidth}
    \centering
    \includegraphics[width=\textwidth]{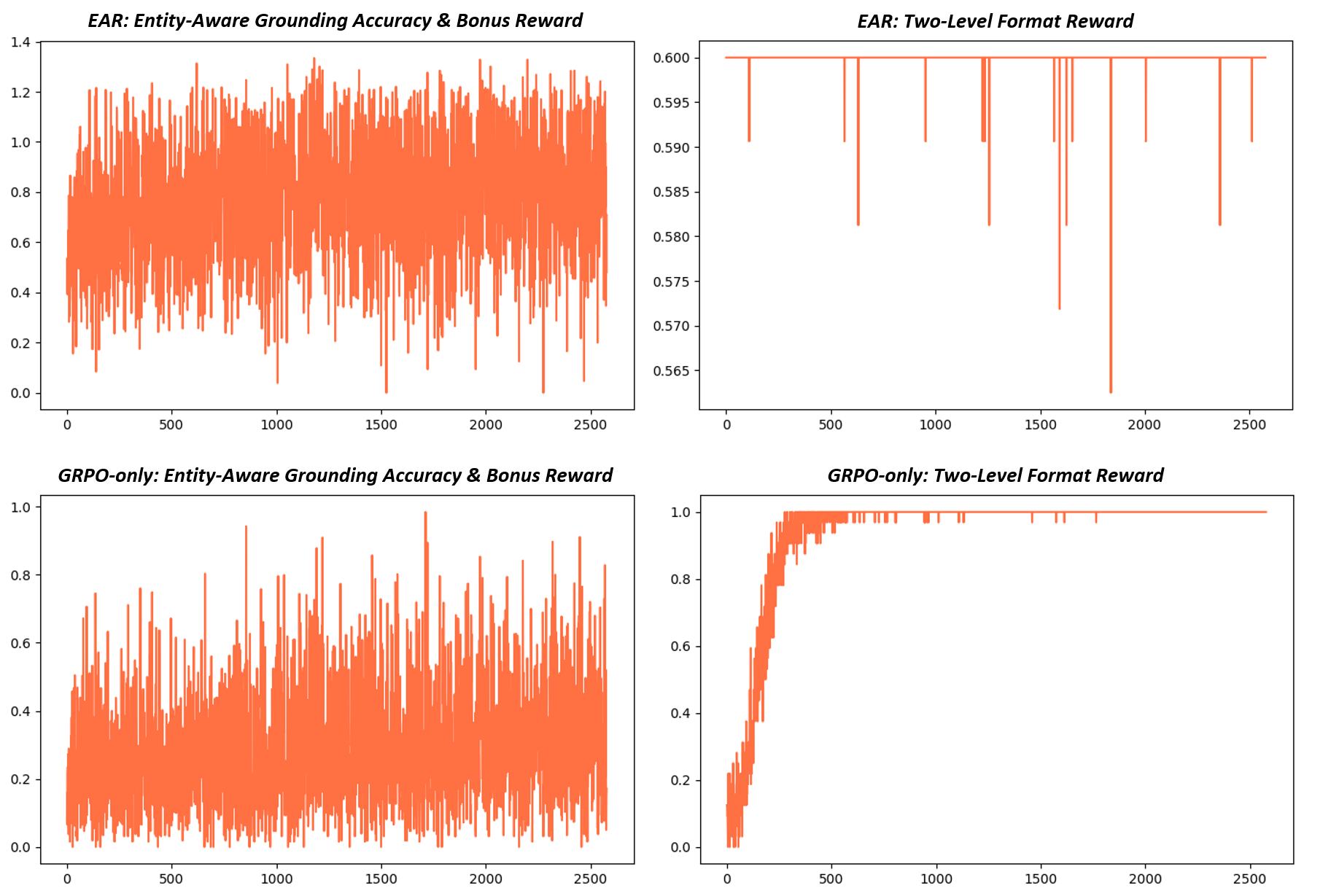}
    \caption{\textbf{Reward Evolution Curves.}}
    \label{Fig6}
  \end{minipage}
\end{figure}

 From the failure cases shown in Fig.~\ref{Fig5}, two typical error patterns can be observed. In some scenes, the model identifies relevant entities but does not strictly satisfy the required spatial relations, showing confusion under compound positional constraints. In other scenes, it is misled by similar and repetitive structures and selects incorrect yet plausible regions. These results indicate that strict relational consistency and fine-grained discrimination still remain challenging in complex multi-entity settings. Specifically, within the reasoning traces, we observe a persistent linguistic bias where the model frequently generates the fixed template: ``The description mentioned two entities.'' This occurs regardless of the actual number of objects present. It suggests that the model tends to overfit to this specific phrasing due to the predominance of ``single-subject-single-object'' instances during the SFT phase.
\par More visualization results can be referred to the supplementary material.
\par \noindent \textbf{Visualization on Reward Evolution.} We conduct this analysis on Qwen2.5-VL-3B. As shown in Fig.~\ref{Fig6}, we can analyze in the following points. (1) The relational consistency reward (bonus) and entity-aware grounding accuracy reward are merged for visualization. This is because only a small portion of batches contain multiple objects, making the second term of $R_{rel}$ is sparsely activated (Eq.~\ref{Eq3}). Meanwhile, these two rewards are tightly coupled in calculation in practical implementation. (2) For the GRPO-only setting, the weights of both format levels are increased by 0.2 (i.e. $\lambda_{1} = \lambda_{2} = 0.5$), Without SFT initialization, strict format regularization becomes crucial to prevent structural collapse and stabilize optimization. (3) Compared to GRPO-only mode, EAR exhibits a healthier reward evolution. With SFT anchoring the output structure in advance, the model avoids format collapse and reduces fluctuation during optimization. This structural stabilization enhances the model’s focus on entity grounding and relational consistency, which results in a higher average reward throughout training.
\section{Conclusion}
In this paper, we introduce ME-RSRG, a challenging dataset for multi-entity remote sensing reasoning grounding. Unlike conventional grounding settings that emphasize isolated entities, ME-RSRG explicitly requires role identification, relational reasoning, and coordinated multi-entity localization. To address this challenge and build a benchmark for this task, we further propose an Entity-Aware Reasoning (EAR) framework. EAR introduces a tailored Entity-Aware Reward-Driven GRPO strategy and adopts a two-stage optimization scheme. With this design, it stabilizes reasoning behavior and strengthens multi-entity relational grounding capability. Extensive experiments demonstrate that ME-RSRG poses substantial difficulty for existing models, while EAR achieves consistent improvements across multiple visual-linguistic foundation models.

\section*{Acknowledgements}
This work was supported by ``the Fundamental Research Funds for the Central Universities'' (Grant number 501QYJC2025102002)


%
%
\bibliographystyle{splncs04}
\bibliography{main}

@String(CVPR  = {IEEE Conf. Comput. Vis. Pattern Recog.})

@String(ICCV  = {Int. Conf. Comput. Vis.})

@String(ECCV  = {Eur. Conf. Comput. Vis.})

@String(NeurIPS = {Adv. Neural Inform. Process. Syst.})

@String(ICLR  = {Int. Conf. Learn. Represent.})

@String(IJCAI = {IJCAI})

@String(ACMMM = {ACM Int. Conf. Multimedia})

@String(TGRS  = {IEEE Trans. Geosci. Remote. Sens.})

@String(CVPR  = {CVPR})

@String(ICCV  = {ICCV})

@String(ECCV  = {ECCV})

@String(NeurIPS = {NeurIPS})

@String(ICLR  = {ICLR})

@String(ACMMM = {ACM MM})

@String(TRGS    = {TGRS})

@article{OpenAI-o1,
      title={OpenAI o1 System Card}, 
      author={Aaron Jaech and Adam Kalai and Adam Lerer and others},
      year={2024},
      journal={arXiv preprint arXiv:2412.16720},
      archivePrefix={arXiv},
}

@article{DeepSeek-R1,
      title={DeepSeek-R1: Incentivizing Reasoning Capability in LLMs via Reinforcement Learning}, 
      author={Daya Guo and Dejian Yang and Haowei Zhang and others},
      year={2025},
      journal={arXiv preprint arXiv:2501.12948},
      archivePrefix={arXiv},
}

@article{Kimi,
      title={Kimi k1.5: Scaling Reinforcement Learning with LLMs}, 
      author={Angang Du and Bofei Gao and Bowei Xing and others},
      year={2025},
      journal={arXiv preprint arXiv:2501.12599},
      archivePrefix={arXiv},
}

@InProceedings{Visual-RFT,
    author    = {Liu, Ziyu and Sun, Zeyi and Zang, Yuhang and others},
    title     = {Visual-RFT: Visual Reinforcement Fine-Tuning},
    booktitle = {ICCV},
    year      = {2025},
    pages     = {2034-2044}
}

@inproceedings{Reason-RFT,
   title={Reason-{RFT}: Reinforcement Fine-Tuning for Visual Reasoning of Vision Language Models},
   author={Huajie Tan and Yuheng Ji and Xiaoshuai Hao and others},
   booktitle={NeurIPS},
   year={2025},
}

@inproceedings{RSVG,
  author       = {Yuxi Sun and
                  Shanshan Feng and
                  Xutao Li and others},
  title        = {Visual Grounding in Remote Sensing Images},
  booktitle    = {ACMMM},
  pages        = {404--412},
  year         = {2022},
}

@article{DIOR-RSVG,
  author       = {Yang Zhan and
                  Zhitong Xiong and
                  Yuan Yuan},
  title        = {{RSVG:} Exploring Data and Models for Visual Grounding on Remote Sensing Data},
  journal      = {TGRS},
  volume       = {61},
  pages        = {1--13},
  year         = {2023},
}

@article{RSVG-HR,
  title={Language query-based transformer with multiscale cross-modal alignment for visual grounding on remote sensing images},
  author={Lan, Meng and Rong, Fu and Jiao, Hongzan and others},
  journal={TGRS},
  volume={62},
  pages={1--13},
  year={2024},
}

@inproceedings{LGFormer,
  author       = {Biao Liu and
                  Xu Liu and
                  Lingling Li and others},
  title        = {Language-Guided Hybrid Representation Learning for Visual Grounding on Remote Sensing Images},
  booktitle    = {IJCAI},
  pages        = {1557--1566},
  year         = {2025},
}

@article{CSDNet,
  author       = {Yichen Zhao and
                  Yaxiong Chen and
                  Ruilin Yao and others},
  title        = {Context-driven and sparse decoding for Remote Sensing Visual Grounding},
  journal      = {Information Fusion},
  volume       = {123},
  pages        = {103296},
  year         = {2025},
}

@article{GRPO,
      title={DeepSeekMath: Pushing the Limits of Mathematical Reasoning in Open Language Models}, 
      author={Zhihong Shao and Peiyi Wang and Qihao Zhu and others},
      year={2024},
      journal={arXiv preprint arXiv:2402.03300},
      archivePrefix={arXiv},
}

@inproceedings{Grounding-Dino,
  author       = {Shilong Liu and
                  Zhaoyang Zeng and
                  Tianhe Ren and others},
  title        = {Grounding {DINO:} Marrying {DINO} with Grounded Pre-training for Open-Set Object Detection},
  booktitle    = {ECCV},
  volume       = {15105},
  pages        = {38--55},
  year         = {2024},
}

@inproceedings{RefCOCOg,
  author       = {Junhua Mao and
                  Jonathan Huang and
                  Alexander Toshev and others},
  title        = {Generation and Comprehension of Unambiguous Object Descriptions},
  booktitle    = {CVPR},
  pages        = {11--20},
  year         = {2016},
}

@inproceedings{RefCOCO,
  title={Modeling context in referring expressions},
  author={Yu, Licheng and Poirson, Patrick and Yang, Shan and others},
  booktitle={ECCV},
  volume = {9906},
  pages={69--85},
  year={2016},
}

@inproceedings{Referitgame,
  title={Referitgame: Referring to objects in photographs of natural scenes},
  author={Kazemzadeh, Sahar and Ordonez, Vicente and Matten, Mark and others},
  booktitle={EMNLP},
  pages={787--798},
  year={2014}
}

@article{TransVG++,
  title={Transvg++: End-to-end visual grounding with language conditioned vision transformer},
  author={Deng, Jiajun and Yang, Zhengyuan and Liu, Daqing and others},
  journal={TPAMI},
  volume={45},
  number={11},
  pages={13636--13652},
  year={2023}
}

@article{Effi-Grounding-Dino,
  author       = {Zibo Hu and
                  Kun Gao and
                  Xiaodian Zhang and others},
  title        = {Efficient Grounding {DINO:} Efficient Cross-Modality Fusion and Efficient Label Assignment for Visual Grounding in Remote Sensing},
  journal      = {TGRS},
  volume       = {63},
  pages        = {1--14},
  year         = {2025},
}

@inproceedings{Fact-RLHF,
    title = "Aligning Large Multimodal Models with Factually Augmented {RLHF}",
    author = "Sun, Zhiqing  and
      Shen, Sheng  and
      Cao, Shengcao  and others",
    booktitle = "ACL (Findings)",
    year = "2024",
    pages = "13088--13110",
}

@INPROCEEDINGS{RLHF-V,
  author={Yu, Tianyu and Yao, Yuan and Zhang, Haoye and others},
  booktitle={CVPR}, 
  title={RLHF-V: Towards Trustworthy MLLMs via Behavior Alignment from Fine-Grained Correctional Human Feedback}, 
  year={2024},
  volume={},
  number={},
  pages={13807-13816},}

@inproceedings{POVID,
title={Aligning Modalities in Vision Large Language Models via Preference Fine-tuning},
author={Yiyang Zhou and Chenhang Cui and Rafael Rafailov and others},
booktitle={ICLR},
year={2024},
}

@article{PPO,
      title={Proximal Policy Optimization Algorithms}, 
      author={John Schulman and Filip Wolski and Prafulla Dhariwal and others},
      year={2017},
      journal={arXiv preprint arXiv:1707.06347},
      archivePrefix={arXiv},
}

@inproceedings{DPO,
  author       = {Rafael Rafailov and
                  Archit Sharma and
                  Eric Mitchell and others},
  title        = {Direct Preference Optimization: Your Language Model is Secretly a Reward Model},
  booktitle    = {NeurIPS},
  volumn = {36},
  pages = {53728-53741},
  year         = {2023},
}

@article{So-Fake,
      title={So-Fake: Benchmarking and Explaining Social Media Image Forgery Detection}, 
      author={Zhenglin Huang and Tianxiao Li and Xiangtai Li and others},
      year={2025},
      journal={arXiv preprint arXiv:2505.18660},
      archivePrefix={arXiv},
}

@inproceedings{Vision-R1,
title={Vision-R1: Incentivizing Reasoning Capability in Multimodal Large Language Models},
author={Wenxuan Huang and Bohan Jia and Zijie Zhai and others},
booktitle={ICLR},
year={2026},
}

@article{SegEarth-R1,
      title={SegEarth-R1: Geospatial Pixel Reasoning via Large Language Model}, 
      author={Kaiyu Li and Zepeng Xin and Li Pang and others},
      year={2025},
      journal={arXiv preprint arXiv:2504.09644},
      archivePrefix={arXiv},
}

@article{RemoteReasoner,
  title={RemoteReasoner: Towards Unifying Geospatial Reasoning Workflow},
  author={Yao, Liang and Liu, Fan and Lu, Hongbo and others},
  journal={arXiv preprint arXiv:2507.19280},
  year={2025},
  archivePrefix={arXiv},
}

@INPROCEEDINGS{SkySense-O,
  author={Zhu, Qi and Lao, Jiangwei and Ji, Deyi and others},
  booktitle={CVPR}, 
  title={SkySense-O: Towards Open-World Remote Sensing Interpretation with Vision-Centric Visual-Language Modeling}, 
  year={2025},
  volume={},
  number={},
  pages={14733-14744},}

@article{RSThinker,
  title={Towards Faithful Reasoning in Remote Sensing: A Perceptually-Grounded GeoSpatial Chain-of-Thought for Vision-Language Models},
  author={Jiaqi Liu and Lang Sun and Ronghao Fu and others},
  journal={arXiv preprint arXiv:2509.22221},
  year={2026},
  archivePrefix={arXiv},
}

@article{opt-RSVG,
  title={Language-guided progressive attention for visual grounding in remote sensing images},
  author={Li, Ke and Wang, Di and Xu, Haojie and others},
  journal={TRGS},
  volume={62},
  pages={1--13},
  year={2024},
}

@article{DIOR,
  title={Object detection in optical remote sensing images: A survey and a new benchmark},
  author={Li, Ke and Wan, Gang and Cheng, Gong and others},
  journal={ISPRS},
  volume={159},
  pages={296--307},
  year={2020},
}

@article{Gemini,
  title={Gemini 2.5: Pushing the Frontier with Advanced Reasoning, Multimodality, Long Context, and Next Generation Agentic Capabilities},
  author={Gheorghe Comanici and Eric Bieber and Mike Schaekermann and others},
  journal={arXiv preprint arXiv:2507.06261},
  year={2025},
  archivePrefix={arXiv},
}

@article{HRRSD,
  author       = {Yuanlin Zhang and
                  Yuan Yuan and
                  Yachuang Feng and others},
  title        = {Hierarchical and Robust Convolutional Neural Network for Very High-Resolution Remote Sensing Object Detection},
  journal      = {TGRS},
  volume       = {57},
  number       = {8},
  pages        = {5535--5548},
  year         = {2019},
}

@inproceedings{VRSBench,
 author = {Li, Xiang and Ding, Jian and Elhoseiny, Mohamed},
 booktitle = {NeurIPS},
 pages = {3229--3242},
 title = {VRSBench: A Versatile Vision-Language Benchmark Dataset for Remote Sensing Image Understanding},
 volume = {37},
 year = {2024}
}

@ARTICLE{DOTA-v2,
  author={Ding, Jian and Xue, Nan and Xia, Gui-Song and others},
  journal={TPAMI}, 
  title={Object Detection in Aerial Images: A Large-Scale Benchmark and Challenges}, 
  year={2022},
  volume={44},
  number={11},
  pages={7778-7796}}

@article{Qwen2.5-VL,
  title={Qwen2.5-VL Technical Report},
  author={Shuai Bai and Keqin Chen and Xuejing Liu and others},
  journal={arXiv preprint arXiv:2502.13923},
  year={2025},
  archivePrefix={arXiv},
}

@article{Qwen3-VL,
  title={Qwen3-VL Technical Report},
  author={Shuai Bai and Yuxuan Cai and Ruizhe Chen and others},
  journal={arXiv preprint arXiv:2511.21631},
  year={2025},
  archivePrefix={arXiv},
}

@article{Qwen2-VL,
  title={Qwen2-VL: Enhancing Vision-Language Model's Perception of the World at Any Resolution},
  author={Peng Wang and Shuai Bai and Sinan Tan and others},
  journal={arXiv preprint arXiv:2409.12191},
  year={2024},
  archivePrefix={arXiv},
}

@article{InternVL-2.5,
  title={Expanding Performance Boundaries of Open-Source Multimodal Models with Model, Data, and Test-Time Scaling},
  author={Zhe Chen and Weiyun Wang and Yue Cao and others},
  journal={arXiv preprint arXiv:2412.05271},
  year={2025},
  archivePrefix={arXiv},
}

@article{InternVL-3.5,
  title={InternVL3.5: Advancing Open-Source Multimodal Models in Versatility, Reasoning, and Efficiency},
  author={Weiyun Wang and Zhangwei Gao and Lixin Gu others},
  journal={arXiv preprint arXiv:2508.18265},
  year={2025},
  archivePrefix={arXiv},
}

@article{ms-swift,
  title={SWIFT:A Scalable lightWeight Infrastructure for Fine-Tuning},
  author={Yuze Zhao and Jintao Huang and Jinghan Hu and others},
  journal={arXiv preprint arXiv:2408.05517},
  year={2024},
  archivePrefix={arXiv},
}

@misc{llava,
      title={Visual Instruction Tuning}, 
      author={Liu, Haotian and Li, Chunyuan and Wu, Qingyang and others},
      publisher={NeurIPS},
      year={2023},
      pages = {34892--34916},
      volume = {36},
}

@article{LLaVA-OneVision,
  title={LLaVA-OneVision: Easy Visual Task Transfer},
  author={Bo Li and Yuanhan Zhang and Dong Guo and others},
  journal={arXiv preprint arXiv:2408.03326},
  year={2024},
  archivePrefix={arXiv},
}

@misc{SPCD,
      title={Swimming Pool and Car Detection}, 
      author={Kartik Bhartiya},
      url={https://www.kaggle.com/datasets/kbhartiya83/swimming-pool-and-car-detection},
      year={2019},
}

@inproceedings{ViT,
  author    = {Alexey Dosovitskiy and
               Lucas Beyer and
               Alexander Kolesnikov and others},
  title     = {An Image is Worth 16x16 Words: Transformers for Image Recognition at Scale},
  booktitle = {ICLR},
  year      = {2021},
}

\newpage
\appendix
\section{Prompt Design for Dataset Construction}
\label{section:1}
\begin{figure}[!t]
  \centering
\includegraphics[width=0.96\textwidth]{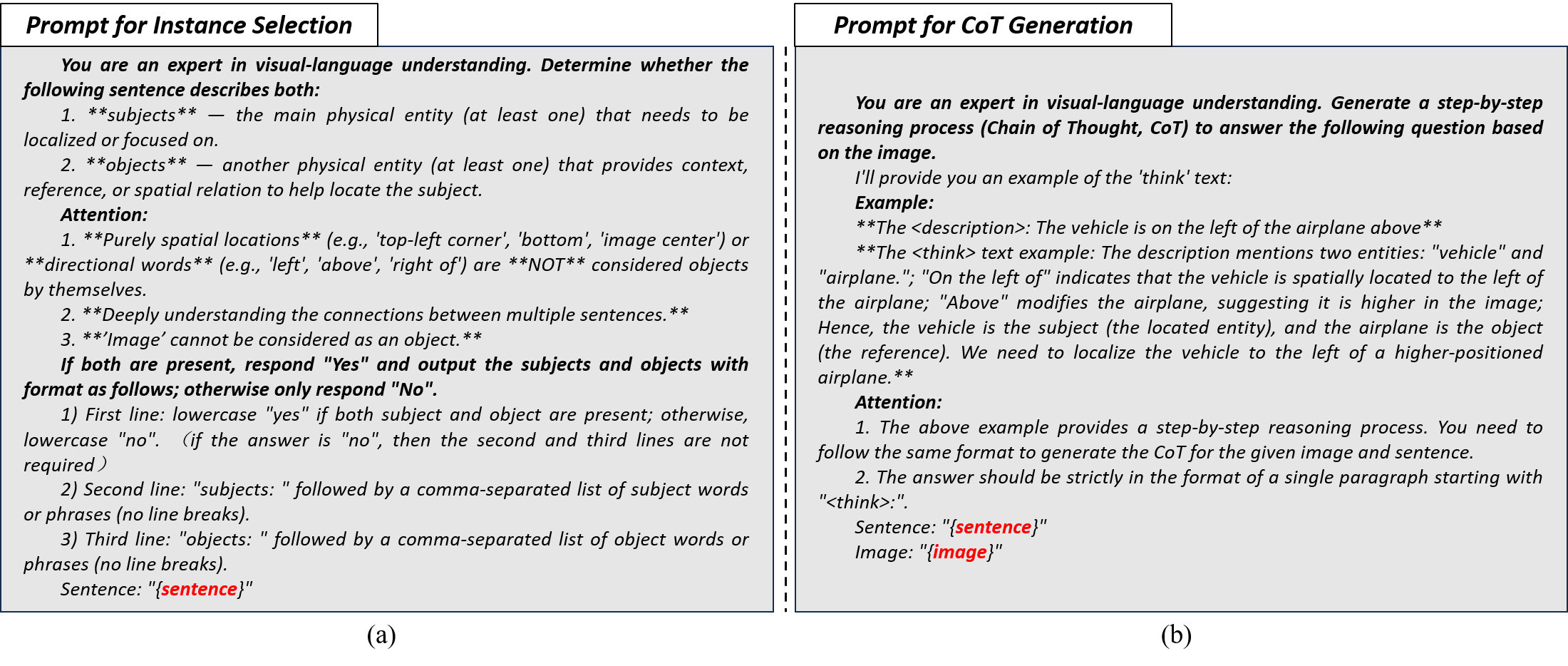}
  \caption{\textbf{Prompt design for instance selection and CoT generation.} Red words indicate the required inputs for the two steps. Best viewed in color.}
  \label{Fig1_supp}
\end{figure}
This section describes the prompt design used in the construction of the ME-RSRG dataset. Specifically, we introduce the prompts for instance selection and for Chain-of-Thought (CoT) generation, which support the filtering of suitable samples and the production of structured reasoning annotations. The prompts designed for these two steps are shown in Fig.~\ref{Fig1_supp}.
\vspace{-0.1cm}
\subsection{Prompt for Instance Selection}
As mentioned in main text, we perform an instance selection step using Gemini-2.5-Pro~\cite{Gemini} to identify suitable samples for multi-entity reasoning. Given each candidate sentence, the model determines whether it contains both a subject entity to be localized and an object entity providing contextual or relational reference. Only sentences that explicitly include both roles are retained for dataset construction. The prompt for this process is provided in Fig.~\ref{Fig1_supp} (a). 
\vspace{-0.1cm}
\subsection{Prompt for CoT Generation}
For reasoning annotation, we also employ Gemini-2.5-Pro~\cite{Gemini} to generate chain-of-thought (CoT) descriptions. Given the image and the corresponding sentence, the model is prompted to produce a step-by-step reasoning process that identifies the subject and object entities and explains their spatial relation for grounding. The generated reasoning is formatted within a <think> paragraph to maintain a consistent structure across the dataset. The prompt is shown in Fig.~\ref{Fig1_supp} (b).
\section{Preliminary Knowledge for Two-Stage Optimization}
\label{section:2}
\subsection{Stage I: Supervised Fine-Tuning (SFT)}
SFT is commonly used to provide an initial capability for structured reasoning and task-specific outputs in visual-linguistic models. Building upon this initialization, reinforcement learning can further refine model behavior by optimizing task-oriented objectives. In this section, we introduce the preliminary knowledge of the two-stage optimization used in our framework, including SFT-based initialization and the subsequent Group Relative Policy Optimization (GRPO). 
\par SFT serves as the initial training stage, where the model learns task-specific output formats and reasoning behaviors from annotated data. Given the constructed dataset, the training objective maximizes the log-likelihood of the target output $o_{i}$ under a standard auto-regressive formulation. The formulation is shown in Eq.~\ref{Eq1}.
\begin{equation}
    \mathcal{L}_{SFT} = - \sum_{t=1}^{|o_{i}|} log \ \pi_{\theta}(o_{i,t}|x_{i}, o_{i,<t})
\label{Eq1}
\end{equation}
where $x_{i}$ denotes the input (``image-text'' pair). $o_{i}$ denotes the output sequence. $o_{i,t}$ is the $t^{th}$ token of $o_{i}$. $\pi_{\theta}$ represents the policy model parameterized by $\theta$.
\subsection{Stage II: Group Relative Policy Optimization (GRPO)}
\par Based on the SFT initialization, we employ GRPO to further optimize the policy with reward signals. For each input $x_{i}$, the policy model $\pi_{\theta}$ samples a set of candidate outputs $\{o_{i}\}_{i=1}^{G}$, which are evaluated by the reward function to compute token-level advantages. The GRPO objective is defined in Eq.~\ref{Eq2}.

\begin{scriptsize}
\begin{equation}
\begin{split}
    \mathcal{L}_{GRPO} &= - \mathbb{E}_{x_{i} \sim D, \{o_{i}\}_{i=1}^{G} \sim \ \pi_{\theta_{old}}(\cdot|x_{i})} \\ & \frac{1}{G}\sum_{i=1}^{G}\frac{1}{|o_{i}|}\sum_{t=1}^{|o_{i}|}min(r_{i,t}(\theta)\hat{A}_{i,t}, clip(r_{i,t}(\theta), 1-\epsilon, 1+\epsilon)\hat{A}_{i,t}) - \beta\mathbb{D}_{KL}(\pi_{\theta}||\pi_{ref}), \\ & \hat{A}_{i,t} = \frac{R_{i}-mean(R)}{std(R)}, \quad r_{i,t}(\theta) = \frac{\pi_{\theta}(o_{i,t}|x_{i}, o_{i,<t})}{\pi_{old}(o_{i,t}|x_{i}, o_{i,<t})}
\end{split}
\label{Eq2}
\end{equation}
\end{scriptsize}
where $R_{i}$ represents the reward assigned to completion $o_{i}$. $\hat{A}_{i,t}$ denotes the normalized advantage computed from group rewards. $r_{i,t}(\theta)$ is the probability ratio between the current policy and the old policy ($\pi_{\theta_{old}}$). $\epsilon$ controls the clipping range, and $\beta\mathbb{D}_{KL}(\pi_{\theta}||\pi_{ref})$ regularizes the policy toward the reference policy.
\begin{table}[!t]
  \caption{\textbf{Optimization details for SFT and GRPO.}}
  \centering
  \scalebox{0.95}{
  \begin{tabular}{c c | c c}
  \cmidrule(r){1-4}
  \multicolumn{2}{c}{\textbf{SFT}} & \multicolumn{2}{c}{\textbf{GRPO}} \\
  \cmidrule(r){1-4}
  Parameters & Values & Parameters & Values \\
  \cmidrule(r){1-4}
  Learning rate & 1e-4 & Learning rate & 1e-5 \\ 
  \cmidrule(r){1-4}
  Warmup ratio & 0.05 & Warmup ratio & 0.01 \\ \cmidrule(r){1-4}
  Max model length & 2048 & Max model length & 2048 \\
  \cmidrule(r){1-4}
  LoRA rank & 8 & LoRA rank & 8 \\ 
  \cmidrule(r){1-4}
  LoRA $\alpha$ & 32 & LoRA $\alpha$ & 32 \\ 
  \cmidrule(r){1-4}
  Gradient accumulation steps & 16 & Gradient accumulation steps & 4 \\
  \cmidrule(r){1-4} 
  - & - & Generations & 8 \\
  \cmidrule(r){1-4}
  - & - & $\beta$ for KL & 0.0025 \\
  \cmidrule(r){1-4}
  \end{tabular}}
 \label{Tab1}
\end{table}
\section{Comprehensive Implementation Details}
\label{section:3}
\subsection{Large Visual-Linguistic Models}
As mentioned in main text, we mainly select the notable Qwen-VL series~\cite{Qwen2-VL, Qwen2.5-VL, Qwen3-VL} and InternVL series~\cite{InternVL-2.5, InternVL-3.5} as the policy models due to their strong multimodal reasoning ability and diverse model scales. LLaVA~\cite{llava} and LLaVA-OneVision~\cite{LLaVA-OneVision} are also selected for a more comprehensive evaluation across different baselines.
\par \noindent \textbf{Qwen-VL Series.} Qwen2-VL pioneers native dynamic resolution, breaking the ``fixed-grid'' limitation to process any aspect ratio. By introducing mRoPE (Multimodal Rotary Position Embedding), it establishes the first robust link between 2D spatial layouts and 1D video timelines. Expanding on Qwen2-VL, Qwen2.5-VL shifts toward efficiency and ``time-aware'' perception by integrating window attention into the ViT~\cite{ViT}. Absolute time encoding further enables precise frame-to-timestamp mapping, extending stable video perception beyond one hour. Qwen3-VL elevates Qwen2.5-VL from perception to native reasoning through visual Chain-of-Thought (CoT). Refined through Deep GRPO~\cite{GRPO}, it masters complex spatial logic for domains like remote sensing. For our experiments, we select Qwen2-VL (2B), Qwen2.5-VL (3B/7B), and Qwen3-VL (4B/8B).
\par \noindent \textbf{InternVL Series.} InternVL is a large-scale visual-linguistic series designed to bridge the gap between open-source models and proprietary systems. In this paper, we select InternVL2.5-4B, InternVL3.5-4B, and InternVL3.5-8B for evaluation. InternVL2.5 series enhances multi-image understanding in logical reasoning, mathematics, and code generation. These optimizations deliver a more efficient balance between computational cost and multimodal intelligence. InternVL3.5 supports higher resolution inputs and utilizes advanced training strategies to minimize hallucinations in dense environments. This version also excels at pixel-level detail and complex spatial layouts.
\par \noindent \textbf{Other Visual-Linguistic Foundation Models.} We also include LLaVA and LLaVA-OneVision in our experiments. These models are widely used open-source visual-linguistic baselines and exhibit strong visual instruction-following capability. Their architectures and training strategies differ from those of the Qwen-VL and InternVL series, providing complementary backbones for evaluation. Additional models are not included mainly due to limited open-source availability, compatibility with our experimental setting, and training resource constraints.
\subsection{Optimization Details}
For the two-stage optimization strategy consisting of SFT and reward-driven GRPO, we provide detailed parameters in Tab.~\ref{Tab1}. Most training parameters are kept consistent across different policy models, while minor adjustments can be found in the released code.
\begin{figure}[!t]
  \centering
\includegraphics[width=0.84\textwidth]{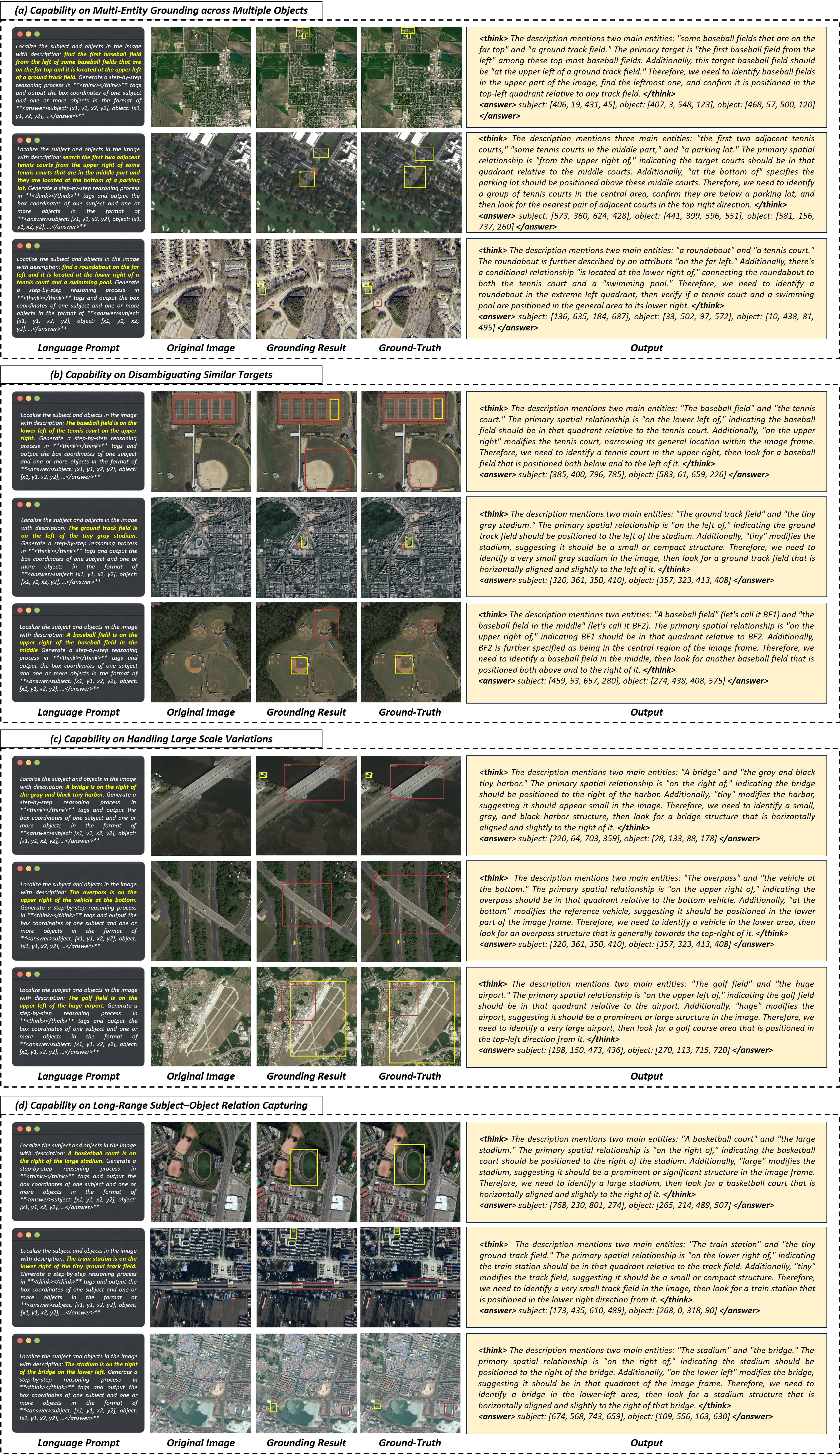}
  \caption{\textbf{Supplementary qualitative visualization results.}}
  \label{Fig2_supp}
\end{figure}
\par SFT is trained with a learning rate of 1e-4 and a warmup ratio of 0.05, with the maximum model length set to 2048. GRPO uses a smaller learning rate of 1e-5 and a warmup ratio of 0.01, with a maximum model length of 2048. Gradient accumulation steps are set to 16 for SFT and 4 for GRPO. During GRPO training, 8 generations are sampled for each query, and the KL regularization coefficient $\beta$ is set to 0.0025. For both stages, LoRA is adopted for with rank 8 and $\alpha=32$. Input images are not resized to a fixed resolution. Instead, the height and width are adjusted to the nearest multiples of 28.
\subsection{Visualization on Qualitative Results.}
\par \noindent \textbf{Capability on Multi-Entity Grounding across Multiple Objects.} As shown in Fig.~\ref{Fig2_supp} (a), the model optimized in EAR framework can accurately identify and grounds the subject among multiple objects while correctly locating the related objects. 
\par \noindent \textbf{Capability on Disambiguating Similar Targets.} As shown in Fig.~\ref{Fig2_supp} (b), the Qwen2.5-VL-7B accurately grounds targets even when multiple similar objects appear in close proximity. This ability reduces confusion among visually similar instances and enables precise subject-object localization. 
\par \noindent \textbf{Capability on Handling Large Scale Variations.} As shown in Fig.~\ref{Fig2_supp} (c), the model effectively grounds entities with large scale differences. It can accurately localize both large and small objects within the same scene, demonstrating strong robustness to scale variation. 
\par \noindent \textbf{Capability on Long-Range Subject-Object Relation Capturing.} As shown in Fig.~\ref{Fig2_supp} (d), Optimized in EAR, visual-linguistic model can capture subject-object relations even when the two entities are far apart in the scene. It correctly associates the distant entities and grounds them according to their relations.
\subsection{Visualization on Failure Cases.} This section presents additional failure cases to analyze the limitations of the proposed EAR framework on the ME-RSRG dataset. 
\par
\begin{figure}[!t]
  \centering
\includegraphics[width=0.9\textwidth]{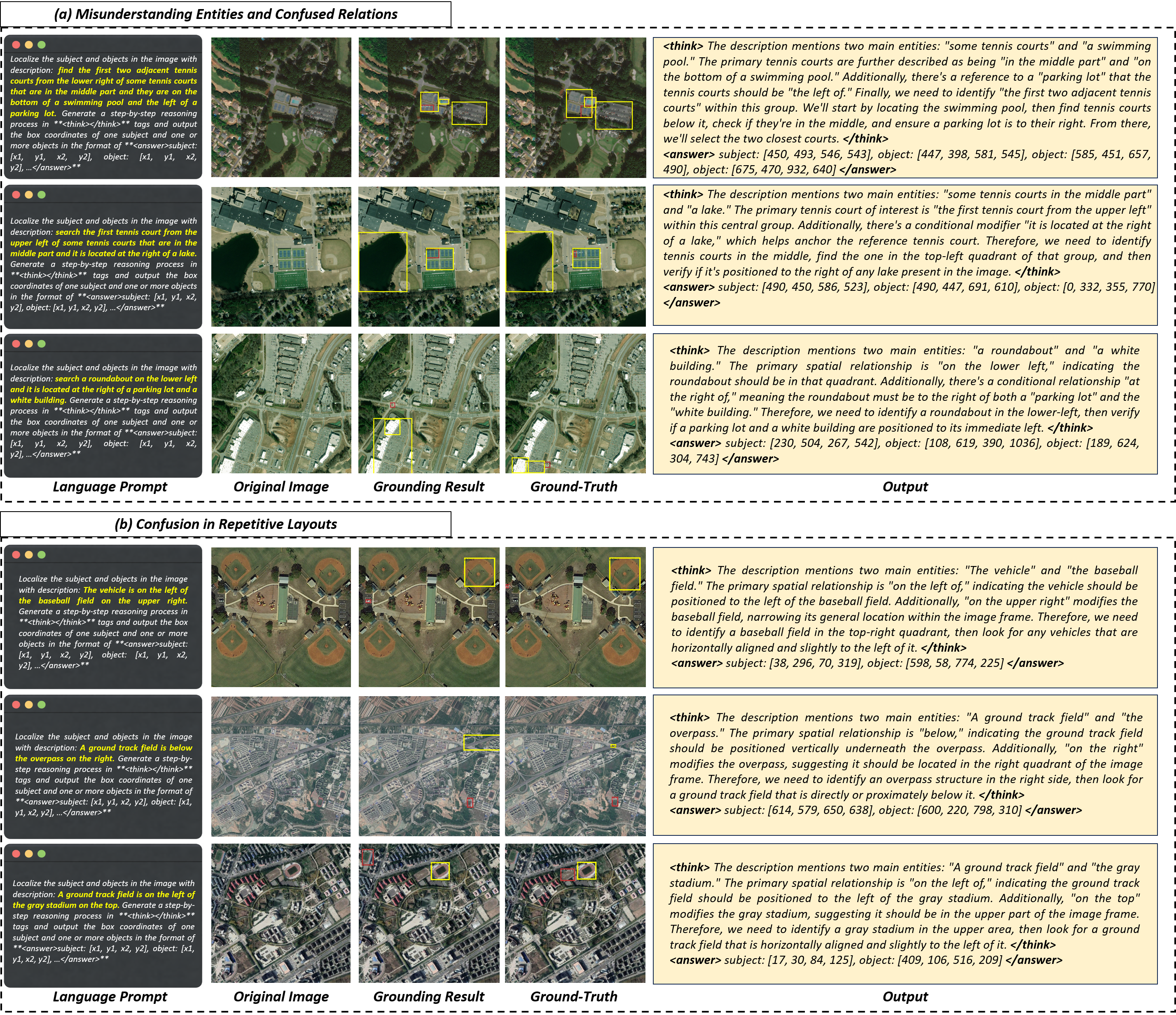}
  \caption{\textbf{Supplementary failure cases visualization.}}
  \label{Fig3_supp}
\end{figure}

 \noindent \textbf{Misunderstanding Entities and Confused Relations.} As shown in Fig.~\ref{Fig3_supp} (a), The model may incorrectly identify entity roles or misunderstand the relation described in the sentence, leading to inaccurate grounding results. For the first case, the object ``some tennis courts that are in the middle part'' is not fully localized, as the bounding box fails to cover the entire region. This incomplete grounding of the object further affects the localization of the subject ``the first two adjacent tennis courts''. For the second case, the object ``some tennis courts'' is visually divided into four groups, each containing three courts. The model interprets ``first'' as the first group rather than the first one, leading to incorrect localization. For the third case, different interpretations of the object (``parking lot'' and ``white building'') lead to completely different subject predictions. This example also highlights the importance of subject modifiers such as ``lower left''.
\par \noindent \textbf{Confusion in Repetitive Layouts.} As shown in Fig.~\ref{Fig3_supp} (b), Although EAR performs well in distinguishing similar targets, highly repetitive layouts can still mislead the model and lead to plausible but incorrect grounding. For the first case, the long distance between entities makes the relationship ``to the left of'' vulnerable to interference from similar vehicles, resulting in incorrect grounding. For the second case, the model fails to locate the relevant object entity due to insufficient visual features. The third case shows a typical phenomenon that the correct spatial cue is overridden by more salient but misleading visual similarities among targets.
\end{document}